\documentclass[
    manuscript,
    screen,
sigconf, language=english]{acmart}

\usepackage{framed}
\usepackage{mdframed}                       
\usepackage{enumitem}                       
\usepackage{wrapfig}                        
\usepackage{lipsum}                         
\usepackage{soul}                           
\usepackage{pdfpages}                       
\usepackage{ragged2e}                       
\usepackage[export]{adjustbox}              

\usepackage[
    nameinlink
]{cleveref}                                 

\crefname{section}{Section}{Sections}
\Crefname{section}{Section}{Sections}
\crefname{figure}{Fig.}{Figs.}
\Crefname{figure}{Fig.}{Figs.}
\crefname{table}{Tab.}{Tabs.}
\Crefname{table}{Tab.}{Tabs.}
\crefname{equation}{Eq.}{Eqs.}
\Crefname{equation}{Eq.}{Eqs.}
\crefname{algorithm}{Alg.}{Algs.}
\Crefname{algorithm}{Alg.}{Algs.}
\crefname{appendix}{Appendix}{Appendices}
\Crefname{appendix}{Appendix}{Appendices}
\crefname{chapter}{Chapter}{Chapters}
\Crefname{chapter}{Chapter}{Chapters}

\DeclareCaptionLabelFormat{andtable}{#1~#2  \&  \tablename~\thetable}

\newcommand{\fig}[1]{Fig.~\ref{#1}}

\newcommand{\ie}{\textit{i.e.}}             
\newcommand{\eg}{\textit{e.g.}}             
\newcommand{\Eg}{\textit{E.g.}}             
\newcommand{\apr}{\textit{a priori}}        
\newcommand{\exan}{\textit{ex-ante}}    
\newcommand{\expo}{\textit{ex-post}}    
\newcommand{\etc}{\textit{etc}.}           

\setlist{leftmargin=*}

\setcopyright{acmcopyright}
\copyrightyear{2023}
\acmYear{2023}
\acmDOI{XXXXXXX.XXXXXXX}

\begin{document}

\title[Reward Reports]{Reward Reports for Reinforcement Learning}

\author{Thomas Krendl Gilbert}
\email{tg299@cornell.edu}
\orcid{0000-0003-1029-4535}
\affiliation{%
    \institution{Digital Life Initiative, Cornell Tech}
    \city{New York}
    \state{New York}
    \country{USA}
}

\author{Nathan Lambert}
\email{nathan@huggingface.co}
\orcid{0000-0002-9997-6817}
\affiliation{%
    \institution{HuggingFace}
    \city{Berkeley}
    \state{California}
    \country{USA}
}

\author{Sarah Dean}
\email{sdean@cornell.edu}
\orcid{XXXX-XXXX-XXXX-XXXX}
\affiliation{%
    \institution{Cornell University}
    \city{Ithaca}
    \state{New York}
    \country{USA}
}

\author{Tom Zick}
\email{tzick@jd24.law.harvard.edu}
\orcid{XXXX-XXXX-XXXX-XXXX}
\affiliation{%
    \institution{Harvard Law School}
    \city{Boston}
    \state{Massachusetts}
    \country{USA}
}

\author{Aaron Snoswell}
\email{a.snoswell@qut.edu.au}
\orcid{0000-0003-0960-0144}
\affiliation{%
    \institution{Centre for Automated Decision-Making and Society, Queensland University of Technology}
    \city{Brisbane}
    \state{Queensland}
    \country{Australia}
}

\renewcommand{\shortauthors}{Gilbert, et al.}

\begin{abstract}
    Building systems that are good for society in the face of complex societal effects requires a dynamic approach. 
Recent approaches to machine learning (ML) documentation have demonstrated the promise of discursive frameworks for deliberation about these complexities. 
However, these developments have been grounded in a static ML paradigm, leaving the role of feedback and post-deployment performance unexamined. 
Meanwhile, recent work in reinforcement learning has shown that the effects of feedback and optimization objectives on system behavior can be wide-ranging and unpredictable. 
In this paper we sketch a framework for documenting deployed and iteratively updated learning systems, which we call \textit{Reward Reports}.
Taking inspiration from various contributions to the technical literature on reinforcement learning, we outline Reward Reports as living documents that track updates to design choices and assumptions behind what a particular automated system is optimizing for. 
They are intended to track dynamic phenomena arising from system deployment, rather than merely static properties of models or data. 
After presenting the elements of a Reward Report, we discuss a concrete example: Meta's BlenderBot 3 chatbot. 
Several others for game-playing (DeepMind's MuZero), content recommendation (MovieLens), and traffic control (Project Flow) are included in the appendix.


\end{abstract}




\maketitle

\section{Introduction}

Algorithmic systems often impact society in profound and difficult to anticipate ways. 
To assess risk, a system designer must take into account not only immediately impacted stakeholders, but also third party externalities and the feedback loops they may engender. 
As AI matures and is deployed in new ways, emerging capabilities challenge what designers and other stakeholders assume algorithmic systems can do, making a priori risk assessment of these ``agents'' even harder~\cite{chan2023harms}. 

The recent rise of dialogue agents powered by Large Language Models (LLMs) is a good example. 
These agents are trained on inconceivably large data corpora, deriving extremely sophisticated linguistic representations. 
More importantly, the form and richness of their conversations with users generate effects that are not reducible to one-off interactions. 
Particular responses cannot be meaningfully isolated from prolonged exchanges; users may be influenced long afterward.
Beyond the biases present in individual outputs, both the dialogue agents themselves and derived data artefacts (e.g. user chat histories) may be integrated into search architectures or other online services that qualitatively alter users' relationship with the web.
As a result, diagnostics or audits of LLMs alone are an insufficient guide for design interventions.
To manage feedback-heavy systems responsibly, the repercussions of algorithmic changes must be reflexively documented.
In other words, both emergent system behaviors and the changing assumptions of key stakeholders about them must be accounted for in an ongoing and responsive manner.

We propose a new form of documentation, \textit{Reward Reports}, to move the research community towards a world where these changes are regularly tracked, reflected upon, and responded to. 
For designers, this documentation would aid internal efforts at reverse engineering the behavior of black box systems, and provide a framework to disentangle complex effects as they manifest. 
Moreover, a standard mechanism for continuous transparency would help harmonize external efforts at domain monitoring and inform regulatory action.
To this end, we are building Reward Reports for popular machine learning systems via community contributions\footnote{Reward Reports are produced and maintained here: \url{https://github.com/RewardReports/reward-reports}.}.

Multiple frameworks for documenting AI systems, datasets and models already exist~\cite{mitchell2019model,gebru2021datasheets, richards2020methodology}. 
However, these approaches all aim to track sources of potential bias or harm within a static machine learning (ML) paradigm.
One might imagine that issuing successive Model Cards would be sufficient to monitor the behavior of deployed systems.
However, system architectures display several key features that would make such a regime insufficient as a basis for accountability. 
First, the effects of deployed AI systems are not static, and the dynamic impacts of successive system updates can subvert efforts both to manage downstream harms and to more evenly distribute benefits to vulnerable subpopulations. Reflecting on these changes explicitly should be a part of deploying AI responsibly.  
Second, Model Cards do not document the designers' recourse to ML itself in the context of system performance--why specific learning algorithms were chosen, what is expected of them, and what evidence would challenge those assumptions. Checking assumptions is a cardinal part of promoting accountability. 
Third, learned task representations often lie behind external interfaces (such as static APIs), access to which is decoupled from how trained models may change over time. Bridging this gap is crucial in order to understand AI systems in context. 
These features transcend existing documentation regimes. 
The presence of diverse feedback profiles and ongoing dynamics suggests unique risk vectors that must be made interpretable through documentation. 
Full accountability requires a cohesive understanding of how the system incorporates different types of feedback: from historical data, from stakeholders, and from a system's own usage once deployed. 
Reward Reports are designed to foreground these elements, allowing better insight into the societal impacts of data-driven optimization systems where feedback effects play a key role. 

As a framing device, Reward Reports utilize Reinforcement learning (RL), a sub-field of ML that is tasked with solving sequential, open-ended problems.
RL provides a dynamical lens that is broadly applicable to many algorithmic systems with repeated data-driven optimizations. 
Critically, this lens is also applicable to `static' ML systems. 
In many of these systems, new  behaviors emerge post-deployment in response to ongoing usage, and as the system is retrained or applied to new populations. 
Building on proposals to document datasets and models, we focus on reward functions: the objective that guides optimization decisions in feedback-laden systems. 
Reward Reports comprise questions that highlight the promised benefits and potential risks entailed in defining what is being optimized in an algorithmic system, whether explicitly or implicitly construed as RL. 
They are intended as living documents that dissolve the distinction between \exan{} specification and \expo{} evaluation. 
As a result, Reward Reports provide a framework for ongoing deliberation and accountability after a system is deployed, ensuring that desired properties persist in the system’s behavior over time.


In \cref{sec:related-work}, we situate Reward Reports within the existing literature on AI documentation and governance mechanisms. 
In \cref{sec:background}, we review optimization in the context of feedback, focusing on the notions of action, objective, and adaptation. 
In \cref{sec:motivation}, we contextualize these elements of optimization within a taxonomy of feedback. 
Finally, in \cref{sec:examples,sec:discussion-and-conclusion}, we provide a list of questions that interrogate specification components and present an example Reward Report for the BlenderBot 3 "AI chatbot" developed and deployed by Meta. 
Throughout, we maintain a running example of a dialogue agent to illustrate the challenges in documenting feedback-laden and data-driven optimization systems, whether or not they explicitly utilize the RL framework.

\section{Related Work}
\label{sec:related-work}

\subsection{Documentation for AI Systems}

There are several existing proposals for AI system documentation, with some frameworks focused on specific aspects of an AI system, while others examine the system as a whole.
Reward Reports are focused on documenting risks of dynamic machine learning systems.
This complements current documentation efforts by explicating the intended performance characteristics in light of system behaviors that emerge over time through feedback effects such as shifts in usage or model re-training.

\paragraph{Data documentation.} 
Documenting data, regardless of resulting systems, is a well explored avenue ~\cite{barclay2019towards,afzal2021data,hutchinson2021towards,denton2021genealogy, gebru2021datasheets}.
These efforts have helped foster discussion on responsible data collection, as well as showcase issues of bias and representation. However, this paradigm is most impactful within the batch or offline setting of static unsupervised or supervised ML. 
Dynamic systems driven by sequential feedback also use data, but not only in the form of static datasets. 
Rather, both RL and deployed ML systems generate and eventually transform data. This is due both to the optimization decisions of the systems themselves and the dynamics of the environment in which they operate. 
Past data documentation efforts are therefore insufficient to reveal the risks and failures 
of dynamic datasets and feedback driven systems.  

\paragraph{Static system documentation.}
There are also proposals for model~\cite{mitchell2019model}, domain~\cite{ramirez2020drec,kuhl2021conduct}, or outcome specific~\cite{sokol2020explainability} forms of ML documentation.
Reward Reports might be a useful supplement to these approaches for several reasons. 
With regards to Model Cards, there is substantial deliberation entailed in mapping between the chosen optimization and resultant behavior in dynamic systems.
For example, designers must consider a range of alternative specifications that were technically feasible pre-deployment, as well as the types of feedback available to help optimize post-deployment. 
These aspects of system design cannot be captured solely by documenting the model. 
Furthermore, unlike domain-specific reporting frameworks, we provide a 
general template that can be applied to any specific application.
Our work has similarities with previous proposals for AI Ethics Sheets~\cite{mohammad2021ethics}, Fact Sheets~\cite{arnold2019factsheets,richards2020methodology}, or Scorecards~\cite{blasch2021multisource}, but uniquely focuses on prompting deliberation about the feedback-driven risks inherent to dynamic systems.

\paragraph{Auditing and Assessment approaches.}
Algorithmic Impact Assessments (AIA) offer a framework for evaluating risks before an AI system is developed or acquired~\cite{reisman2018algorithmic, Selbst2018TheIA}.
AIAs were inspired by environmental impact assessments, which provided one path to regulate industries in which corporate expertise outpaced government capacity.
These frameworks presume an agency-vendor relationship, and focus narrowly on the procurement of automated decision systems.
Meanwhile, many audit mechanisms attempt to confirm either internally or externally whether a given system conforms to a legacy standard or regulation \cite{raji2020closing}.
Reward Reports are intended to supersede these \exan{} concerns, 
engaging instead with the necessarily non-linear and circuitous process of refining the specifications of a feedback system.

\subsection{Societal Risks of dynamic Machine Learning Systems}
The AI ethics community is increasingly acknowledging the important role of feedback and dynamic effects on system behaviours.
While critical and discursive interpretations of feedback are more common in static ML than RL research \cite{matias2022,reader2022models,lucherini2021t},  the technical RL community is also increasingly aware of the limitations of current algorithmic approaches and evaluation paradigms.
For instance, the RL research community has begun to reflect on the unique risks and challenges that may be posed by RL systems, in particular those that leverage black-box neural networks for decision-making.
There are whitepapers charting these challenges 
~\cite{whittlestone2021societal,gilbert2021mapping}, as well as attempts to address societal risks through technical means~\cite{wen2021algorithms,carroll2021estimating,evans2021user,zhan2021towards}.
There is also the recent rise of RL from human feedback (RLHF) as both a critical step in training LLMs and a metaphor for value alignment \cite{bai2022constitutional}.
Recent general audience books have echoed these tensions ~\cite{russell2019human,christian2020alignment}. 
While these efforts have begun to capture the unique stakes in deploying RL systems, there is no consensus on how to chart associated risks. 
We intend Reward Reports to organize these efforts' reflections into an instrument of deliberation and accountability.  


\subsection{AI Governance}
Governance mechanisms can also be used to dictate safer machine learning practices.
As reflected in the growing number of proposed AI governance frameworks~\cite{gasser2017layered,lee2019webuildai,yeung2019ai,cihon2019standards,wirtz2020dark,reddy2020governance}, ML and adjacent communities have increasingly acknowledged socio-technical risks and the need for novel harm mitigation strategies~\cite{selbst2019fairness,andrus2020ai, dean2021axes}.
These frameworks have begun to influence legislation. 
For example, the Canadian government has mandated the  Algorithmic Risk Assessment tool for procurement \cite{canada2019AIATool}, and current draft legislation from the EU commission calls for AI system documentation tailored to forms of risk (\eg{} in Title 3, Ch. 2, Art. 10-14) \cite{eu2021harmonised}.
While these frameworks provide needed prohibitions and protections, they favor interpretive flexibility over specific design decisions, and often assume static AI systems that need strictly \expo{} documentation.
In contrast, Reward Reports would track requisite design decisions, and provide an interface for stakeholders to reflect on the validity of those design choices over time.
This would in essence dissolve the boundary between \exan{} and \expo{} assessment.

\section{Background}
\label{sec:background}
In this section, we reframe the concept of data-driven optimization in terms of dynamical systems.
The type of optimization we describe encompasses the training of large language models, as well as their effects on the world once deployed.
We begin by reinterpreting these dynamics in terms of
action, objective, and adaptation.
This taxonomy emphasizes the \textit{use} and \textit{behavior} of learned models rather than the closed act of developing them.
We then review the RL framework, and recount its broader connections to optimization and machine learning.

\subsection{Action, Objective, Adaptation}

Learned predictive models are the means to some end.
It is the decisions made, or \emph{actions} taken,  that determine the extent to which a model is successful.
For example, congested suburban roadways in the community of Los Gatos, CA, USA are caused directly by the actions of drivers, and indirectly by the actions of routing algorithms that predict a poorly-scaling short-cut path~\cite{judy2018losgatos}. 
Similarly, the optimization of datacenter operations at Google
uses the predictions of trained models to adjust set-points and load distribution -- the predictions serve as a catalyst for action in the real world~\cite{gao2014datacenter}.

Action occurs not only on the basis of predictions, but also towards some \emph{objective}.
The definition of objective is crucial to the resulting behavior.
Identical traffic models will result in different routing suggestions depending on whether the algorithm is optimizing for arrival time or fuel consumption~\cite{NREL}.
Likewise, content interaction models have vastly different impacts when they are used to uprank posts predicted to receive many `likes' compared with those predicted to receive long comments \cite{hagey2021facebook}.

Finally, these optimization systems are often updated based on additional data collected during their operation, making them \textit{adaptive}.
By accounting for the dependence between past decisions, observed data, and current models, systems in effect react to dynamic environments and improve performance over time.
For example, when observed music listening patterns are used as additional data in preference models, music recommendation algorithms can adapt to an individual's evolving tastes~\cite{dhahri2018mood}.
On the other hand, adaptivity can also exacerbate biased data.
For example, predictive policing systems can amplify racial biases in arrests by directing more patrols towards areas with more documented arrests \cite{lum2016predict,predpol}.

\begin{figure}[t]
    \begin{mdframed}
        \justifying
        { \noindent \bf Large Language Models through the RL Lens} \\
        \begin{small}
            \noindent
            We can draw a direct analogy between the MDP setting and the evaluation of language models.
            While language models are not Markovian in an exact sense, the notion of ``state'' can be applied to the \textit{conversation text} at stake in a particular user interaction.
            The ``observations'' of the language model would consist of the \textit{subset of the historic conversation text} that fits within the context window (the ``time horizon'') of the language model.
            Likewise, the ``actions'' taken by a dialogue agent consist of the \textit{token sequences that are generated} to form responses to user queries.
            The ``dynamics'' of this system correspond to the \textit{user responses} to dialogue agent generations - updating the conversation state by moving the sequence of conversation forward.
            Finally, the \textit{performance metrics} (e.g. loss and/or regularization function(s), user 'thumbs up'/'thumbs down' feedback) that shape the behaviour of the language model during pre-training, fine-tuning, and subsequent updates, can be considered as a source of scalar ``reward'' feedback that depends on the specific conversation state at a given point in time.
            This lens is central to the recent surge in the use of reinforcement learning from human feedback (RLHF)~\cite{christiano2017deep} to further fine-tune language models with respect to human values.
        \end{small}
    \end{mdframed}
    \label{fig:ads_is_rl}
\end{figure}

\begin{figure*}
    \centering
    \includegraphics[width=0.32\textwidth]{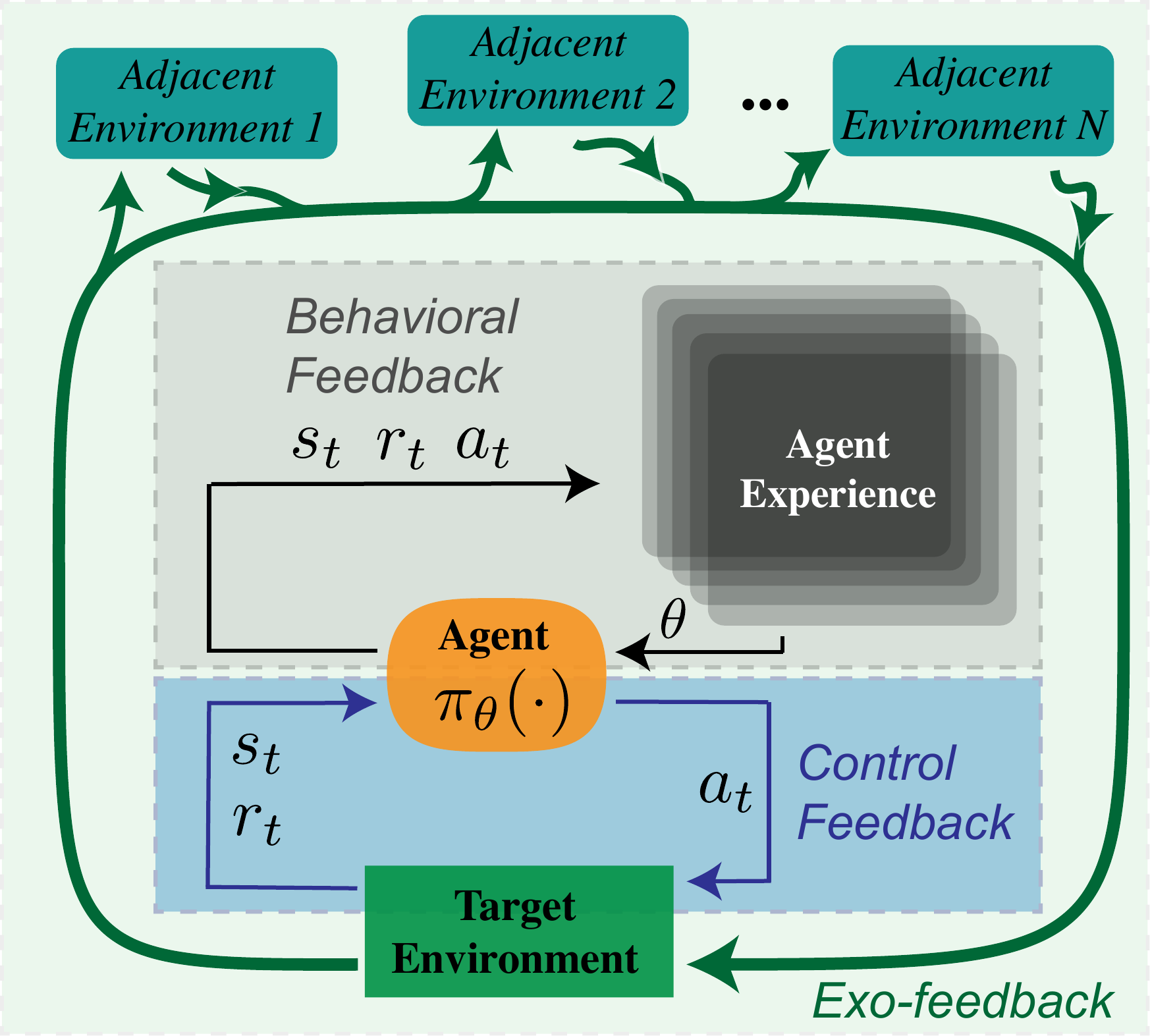}
    \quad
    \resizebox{0.64\textwidth}{!}{
    \renewcommand{\arraystretch}{1.3}
    \begin{tabular}[b]{llll}
    \toprule
    Type of feedback & Feedback Channel   & Dynamics   & Specification Element   \\ \midrule
    Control          & Agent-Environment  & Reaction   & Actions                 \\
    Behavior       & Policy-Reward    & Evaluation & Rewards                 \\
    Exogenous        & Environment-Domain & Drift      & States \& Observations  \\ \bottomrule
    \vspace{20pt}
    \end{tabular}
    }
    \captionlistentry[table]{%
        The relationship between types of feedback, the channel through which information flows, the relationship to dynamics, and specification element(s).
        Different parameters and data interact over time to create dynamic properties internal and external to the RL agent.
    }
    \captionsetup{labelformat=andtable}
    \label{fig:feedback}
    \label{tab:feedback-types}
    \caption{%
        The relationship between types of feedback, the channel through which information flows, the relationship to dynamics, and specification element(s).
        Different parameters and data interact over time to create dynamic properties internal and external to the RL agent.
    }
\end{figure*}

Action, objective, and adaptation are important for ensuring that systems work as intended, even in cases where they are not explicitly defined as part of an underlying model. 
This is especially true for large language models, which act by responding to natural language queries according a variety of engineered objectives: accurate prediction of the next token in a dialogue sequence, but also disparate constraints such as safety, helpfulness, etc. 
Moreover, they already function as parts of a larger adaptive system, as retrained models (GPT-3, GPT-3.5, GPT-4) have been designed and evaluated differently based on their successive integration with the ChatGPT interface.
At present, designers are missing a framework for capturing these properties as dynamic elements of techniques for model optimization and deployment.

\subsection{Reinforcement Learning}

The reinforcement learning (RL) framework succinctly encompasses action, objective, and adaptation. 
RL agents take actions, are motivated by a reward signal which encodes the objective, and adapt based on the feedback from interactions.
While the goal of supervised learning (SL) procedures is to use data to generate a model that makes accurate predictions, the goal of RL algorithms is to
interact with an environment to generate a policy that achieves high reward.
However, once SL models are deployed towards some goal and updated with new data, the concerns highlighted by the RL framework become relevant.
In this sense, ML deployments can be understood through the lens of RL.
This is even more true of language models whose inchoate social effects post-deployment are readily understood within an RL framework -- to say nothing of the explicit use of RL from Human Feedback (RLHF) to fine-tune these models pre-deployment.

In Reinforcement Learning, an \emph{agent} executes \emph{actions} $\vec a_t\in\mathcal A$ in an \emph{environment}.
In response, the agent receives a scalar \emph{reward} $r_t\in\mathbb R$ and makes an \emph{observation} $\vec o_t\in\mathcal O$ of the environment.
Actions are made on the basis of these observations according to a \emph{policy} $\pi:\mathcal H\to\Delta(\mathcal A)$, where $\mathcal H = \mathcal O \times \dots \times \mathcal O$ represents the history of observations and $\Delta(\mathcal A)$ represents a probability distribution over the action space.
The goal of a reinforcement learning agent is to find a policy that maximizes the cumulative reward over some time horizon: $\sum_{t=0}^H \gamma^t r_t$ where the discount factor $\gamma\in(0,1]$ trades-off between immediate and future potential rewards.
As outlined above, this paradigm captures many problems of interest, from choosing advertisements that are most likely to result in a click~\cite{liu2021map,langford2007epoch} to determining the best dosing schedule for a patient~\cite{shortreed2011informing}.

A key element of RL is how actions affect the future behavior of the environment.
This dependence is often modeled as a  Markov Decision Process (MDP)~\cite{bellman1957markovian}.
In the MDP setting, the \emph{state} $\vec s_t$ describes the status of the environment.
The key assumption, called memorylessness, is that the current state and action are sufficient for predicting the future state, \ie{}
\[\mathbb P\{\vec s_{t+1} = \vec s\mid \vec s_0,\dots \vec s_t,\vec a_0,\dots \vec a_t\} = \mathbb P\{\vec s_{t+1} = \vec s\mid \vec s_t, \vec a_t\}\:.\]
The transition probability distribution mapping state-action pairs to subsequent states is referred to as the \emph{system dynamics}.
Furthermore, the scalar reward signal is defined to be determined by the state, so that $r_t = r(\vec s_t, \vec a_t)$ for some reward function $r:\mathcal S\times \mathcal A\to\mathbb R$, mapping from the current environment to a scalar representation of desirability.
Under these assumptions, it is optimal to consider policies that depend only on the current state, $\pi:\mathcal S\to\Delta(\mathcal A)$.
Often, RL algorithms assume that the state is observed directly $\vec o_t=\vec s_t$ (or similarly, that it can be constructed in a straightforward manner \eg{} though history truncation $\vec s_t = [\vec o_{t-h},\dots,\vec o_t]$), and the policy is typically parametric, with a parameter vector denoted $\theta$.


The RL framing is general, so other machine learning paradigms can be viewed as special cases of it as long as key assumptions are named.
For example, supervised learning can be viewed as the optimization of a classification or regression policy where the rewards are defined by accuracy and the time horizon is equal to one.
While, standard supervised learning frameworks do not consider how to update or retrain on the basis of interaction,
there are intermediate points.
Online learning situates supervised learning systems in a sequentially evolving environment~\cite{shalev2011online}, while the study of bandit problems reduces RL to the static regime where actions do not affect the environment~\cite{berry1985bandit}.
The boxed example illustrates how an RL lens can be richly applied to language models as long as terms like \textit{horizon} and \textit{state} are aligned with salient metrics and performance criteria.

\section{Motivation}
\label{sec:motivation}

The RL lens is useful not only because ML deployments often operate in dynamical environments.
It also expounds \emph{feedback} between the environment and the deployment.
In this section, we review three levels of feedback that characterize RL systems, and then motivate Reward Reports as documentation for tracking how and why feedback has been organized.

\subsection{A Taxonomy of Feedback}

We categorize feedback into three types: \textit{control} feedback from state to action, \textit{behavior} feedback from the data to the policy, and \textit{exogenous} (or exo-)feedback from the target environment to adjacent entities~\cite{gilbert2022choices}.
The types of feedback are compared in Tab. 1 and visualized in \fig{fig:feedback}.
Throughout, we continue to illustrate these types of feedback through comparison to a deployed dialogue agent powered by a large language model.

\paragraph{Control feedback}
Control feedback maps observations or states to actions. 
In the case of a dialogue agent, control feedback is the autoregressive language model itself -- a conditional probability distribution from token sequences to subsequent tokens.
``Intelligent'' behaviors arise because actions are constantly adjusted on the basis of observations, even though the rules that control the behavior (the language model distribution) remains the same.

\paragraph{Behavior feedback}
Behaviour feedback maps data from the environment to a learned policy.
This form of feedback occurs when RL systems automatically {adapt} their policy based on {reward}, or when a deployed ML system is updated (e.g. with new weights) to better suit the deployment environment.
Questions of reaction become questions of trial-and-error evaluation: ``What token should follow the preceding tokens'' evolves into ``Can this dialogue agent be made safer/more useful/better aligned with user needs \etc{} through fine-tuning or better prompt engineering'' (for example).
The ability to learn from experience is part of what makes RL systems so powerful, and is what makes RL useful in domains that are difficult to otherwise model.
For example, it would be challenging to hand-design a policy for generating natural language responses to user queries, but data-driven approaches have made this task tractable.

\paragraph{Exogenous feedback}
Exogenous feedback occurs when the application domain itself shifts in response to the deployed system.
These shifts could be due to the system interacting with political or economic conditions that are outside the purview of its formal specification.
Note that the concept of exogenous feedback is richer than the concept of covariate or distribution shift commonly discussed in supervised learning~\cite{koh2021wilds} -- exogenous feedback explicitly foregrounds what lies behind and instigates the shift, i.e. the interactions between the deployed system and the environment.
Dialogue agents are particularly prone to this type of feedback, as their emerging effects on web search, content recommendation, and diverse writing disciplines already demonstrate.
In principle, if such dynamics could be predicted by an RL agent, they could be brought under the purview of behavior or control feedback, but in practice, it is not clear that this is possible -- the observations would need to be sufficiently rich and the planning horizon sufficiently long. 
For these reasons, exogenous feedback highlights the potential of externalized risks in the RL framework.

\subsection{Risks and Documentation}

For many systems, reward design -- the choice of how and what to optimize -- amounts to a political decision about how different types of feedback may rewire the domain and pose risks to various stakeholders.
As it is often impossible to fold all of the domain dynamics within a controllable planning horizon and precise reward function, exo-feedback is fundamentally unavoidable in practice.
Furthermore, it is unrealistic to articulate all possible system specifications \apr{}.
This means that a single specification may simultaneously implicate all three forms of feedback (conrol, behavior, exogenous) outlined above.

The risks of feedback can at least be approached and evaluated through documentation. This calls for legible and periodic mechanisms for auditing RL systems pre- and post-deployment.
It is these reviews that must decide whether or how the optimized behaviors align with the application domain, in correspondence with resultant risks and possible harms.
This may be especially true for dialogue agents whose models may be technically static (insofar as parameters are not updated in real-time in response to user feedback), but whose dynamical effects on society are impossible to specify in advance.
Given the dynamic nature of these effects, the corresponding document must be dynamic as well: updated and revisited over time to map the evolution of feedback between the system and the domain in which it is deployed.

\section{Reward Report Components}
\label{sec:rrs}

\begin{figure}[b]
    \begin{mdframed}[backgroundcolor=white] 
        \begin{center}
            { \Large { \centering { \bf Reward Report Contents}} }
        \end{center}
        
        {\small
        \begin{itemize}
            \item {\bf System Details}: Basic system information.
            \begin{itemize}
                \item Person or organization developing the system
                \item Deployment dates
                \item Contact
            \end{itemize}
            
            \item {\bf Optimization Intent}: The goals of the system and how reinforcement manifests.
            \begin{itemize}
                \item Goal of reinforcement
                \item Performance metrics
                \item Oversight metrics
                \item Failure modes
            \end{itemize}
            
            \item {\bf Institutional Interface}: The interconnections of the automated system with society.
            \begin{itemize}
                \item Involved agencies
                \item Stakeholders
                \item Computation footprint
                \item Explainability
                \item Recourse
            \end{itemize}
            
            \item {\bf Implementation}: The low-level engineering details of the ML system.
            \begin{itemize}
                \item Reward, algorithmic, and environment details
                \item Measurement details
                \item Data flow
                \item Limitations
                \item Engineering artifacts
            \end{itemize}
            
            \item {\bf Evaluation}: Specific audits on system performance.
            \begin{itemize}
                \item Evaluation environment
                \item Offline evaluations
                \item Evaluation validity
                \item Performance standards
            \end{itemize}
            
            \item {\bf System Maintenance}: Plans for long-term verification of behavior.
            \begin{itemize}
                \item Reporting cadence
                \item Update triggers
                \item Changelog
            \end{itemize}
        \end{itemize}
        }
    \end{mdframed}
    \vspace{-1em}
    \caption{Summary of reward report sections and suggested inquiries.}\label{fig:report_sections}
\end{figure}

Here we prescribe Reward Reports, a structured series of design inquiries for automated decision systems (\cref{fig:report_sections}).
Including but not limited to the use of reinforcement learning, Reward Reports are intended to engage practitioners by revisiting design questions over time, drawing reference to previous reports and looking forward to future ones.
As pivotal properties may not be known until the system has been deployed and monitored, the onus is on designers to sustain documentation over time.
This makes Reward Reports into living documents that 
both avoid the limitations of simple, unidirectional answers (\eg{} yes or no) and illuminate societal risks over time.
Moreover, the changelog component of a Reward Report becomes an interface for stakeholders, users, and engineers to continuously oversee and evaluate the documented system.
Thus, Reward Reports are a prerequisite to sociotechnical reflection about the system behavior.

A Reward Report is composed of six sections, arranged to help the reporter understand and document the system. 
A Reward Report begins with \textbf{\textit{system details}} (1) that contain the information context for deploying the model.
From there, the report documents the \textbf{\textit{optimization intent}} (2) which questions the goals of the system and why RL or ML may be a useful tool.
The designer then documents how it can affect different stakeholders in the \textbf{\textit{institutional interface}} (3).
The next two sections contain technical details on the system \textbf{\textit{implementation}} (4) and \textbf{\textit{evaluation}} (5).
The report concludes with plans for \textbf{\textit{system maintenance}} (6) as additional system dynamics are uncovered.



\subsection{System Details}
This section collects basic information a user or stakeholder may need in reference to the automated decision system.
\begin{enumerate}[noitemsep,nolistsep]
\item \textbf{Person or organization deploying the system}: This may be the designer deploying the system, a larger agency or body, or some combination of the two. 
The entity completing the report should also be indicated.
\item \textbf{Reward date(s)}: The known or intended timespan over which this reward function \& optimization is active.
\item \textbf{Feedback \& communication}: Contact information for the designer, team, or larger agency responsible for system deployment.
\item \textbf{Other resources}: Where can users or stakeholders find more information about this system? Is this system based on one or more research papers?
\end{enumerate}

\subsection{Optimization Intent}
This section addresses basic questions about the intent of the reward function and optimization problem. 
Designers first document the intent of a particular solution, translating the system's quantitative objective into a qualitative description.
In later sections, they have the opportunity to further reflect on how implementation details aid in, or diminish the broader goal.
Stakeholders and users can employ this section to understand if the intent of the system matches with the effects they observe or experience.
\begin{enumerate}[noitemsep,nolistsep]
\item \textbf{Goal of reinforcement}: A statement of system scope and purpose, including the planning horizon and justification of a data-driven approach to policy design (\eg{} the use of reinforcement learning or repeated retraining).
This justification should contrast with alternative approaches, like static models and hand-designed policies. 
What is there to gain with the chosen approach?
\item \textbf{Defined performance metrics}: A list of “performance metrics” included explicitly in the reward signal, the criteria for why these metrics were chosen, and from where these criteria were drawn (\eg{} government agencies, domain precedent, GitHub repositories, toy environments). Performance metrics that are used by the designer to tune the system, but not explicitly included in the reward signal should also be reported here.
\item \textbf{Oversight metrics}: Are there any additional metrics not included in the reward signal but relevant for vendor or system oversight (\eg{} performance differences across demographic groups)? Why aren't they part of the reward signal, and why must they be monitored?
\item \textbf{Known failure modes}: A description of any prior known instances of ``reward hacking'' or model misalignment in the domain at stake~\cite{Krakovna2020specification}, and description of how the current system avoids this.
\end{enumerate}

\subsection{Institutional Interface}
This section documents the intended (and in subsequent reports, observed) relationship between the system and the broader context in which it is deployed.
While necessarily piecemeal, the explicit documentation of this interface will allow designers to reflect on and revisit the system assumptions over time. 
These reflections may bring novel interests or agencies into scope and allow for organizing the emergent interests of stakeholders and users where necessary.
\begin{enumerate}[noitemsep,nolistsep]
\item \textbf{Deployment Agency}: What other agency or controlling entity roles, if any, are intended to be subsumed by the RL system? How may these roles change following system deployment?
\item \textbf{Stakeholders}: What other interests are implicated in the design specification or system deployment, beyond the designer? What role will these interests play in subsequent report documentation? What other entities, if any, does the deployed system interface with whose interests are not intended to be in scope?
\item \textbf{Explainability \& Transparency}: Does the system offer explanations of its decisions or actions?
What is the purpose of these explanations?
To what extent is the policy transparent, i.e. can decisions or actions be understood in terms of meaningful intermediate quantities?
\item \textbf{Recourse}: Can stakeholders or users contest the decisions or actions of the system? 
What processes, technical or otherwise, are in place to handle this?

\end{enumerate}

\subsection{Implementation}
Given the sensitivity of reinforcement learning systems, it is important to document specific implementation details of the system.
Even small changes in implementation can result in substantial behavior shifts downstream, making such factors difficult to track when used at scale.
Documenting these design decisions will both help prevent failures in specific applications and assist technical progress.
\begin{enumerate}[noitemsep,nolistsep]
\item \textbf{Reward details}: How was the reward function engineered? \Eg{} is it based on a well-defined metric? Is it tuned to represent a specific behavior? Are multiple terms scaled to make one central loss, and how was the scaling decided?
\item \textbf{Environment details}: Description of states, observations, and actions with reference to planning horizon and hypothesized dynamics/impacts.
What dynamics are brought into the scope of the optimization via feedback? Which dynamics are left external to the system, as drift?
Have there been any observed gaps between conceptualization and resultant dynamics?
\item \textbf{Measurement details}: 
How are the components of the reward and observations measured? 
Are measurement techniques consistent across time and data sources?
Under what conditions are measurements valid and correct?
What biases might arise during the measurement process?
\item \textbf{Algorithmic details}: The key points on the specific algorithm(s) used for learning and planning. 
This includes the form of the policy (e.g. neural network, optimization problem), the class of learning algorithm (e.g. model-based RL, off-policy RL, repeated retraining),
the form of any intermediate model (e.g. of the value function, dynamics function, reward function), technical infrastructure, and any other considerations necessary for implementing the system.  
Is the algorithm publicly documented and is code publicly available? Have different algorithms been used or tried to accomplish the same goal?
\item \textbf{Data flow}: 
How is data collected, stored, and used for (re)training?
How frequently are various components of the system retrained, and 
why was this frequency chosen?
Could the data exhibit sampling bias, and is this accounted for in the learning algorithm?
Is data reweighted, filtered, or discarded? 
Have data sources changed over time? 
\item \textbf{Limitations}: Discussion and justification of modeling choices arising from computational, statistical, and measurement limitations. 
How might (or how have) improvements in computational power and data collection change(d) these considerations and impact(ed) system behavior?
\item \textbf{Engineering tricks}: RL systems are known to be sensitive to implementation tricks that are key to performance. 
Are there any design elements that have a surprisingly strong impact on performance? E.g. state-action normalization, hard-coded curricula, model-initialization, loss bounds, or more?
\end{enumerate}



\subsection{Evaluation}
Assessing the potential behavior of a feedback system is important for anticipating its future performance and risks that may arise.
This section records evaluations done by the designer before deploying the system and each time the reward report is revisited.
This section allows stakeholders and users to hold designers accountable for the performance of the system once deployed.
It is important to distinguish whether the evaluations are done in a simulation (\textit{offline}) or deployed on real users (\textit{online}) and if the evaluation procedure is on a fixed dataset (\textit{static}) or evolves over time (\textit{dynamic}).
\begin{enumerate}[noitemsep,nolistsep]
\item \textbf{Evaluation environment}:
How is the system evaluated (and if applicable, trained) prior to deployment (\eg{} using simulation, static datasets, \etc{})?
Exhaustive details of the {offline evaluation} environment should be provided.
For simulation, details should include description or external reference to the underlying model, ranges of parameters, etc.
For evaluation on static datasets, considering referring to associated documentation (\eg{} \emph{Datasheets}~\cite{gebru2021datasheets}).
\item \textbf{Offline evaluations}: Present and discuss the results of offline evaluation.
{For static evaluation, consider referring to associated documentation (\eg{} \emph{Model Cards}~\cite{mitchell2019model}).}
If applicable, compare the behaviors arising from counterfactual specifications (\eg{} of states, observations, actions).
\item \textbf{Evaluation validity}: To what extent is it reasonable to draw conclusions about the behavior of the deployed system based on the available offline evaluations?

How is the online performance of the system presently understood?
If the system has been deployed, were any unexpected behaviors observed? 
\item \textbf{Performance standards}: What standards of performance and safety is the system required to meet? Where do these standards come from? How is the system verified to meet these standards?
\end{enumerate}


\subsection{System Maintenance}
This section documents plans for post-deployment oversight, including subsequent reviews of real-world implementation and how the monitoring of resultant dynamics is intended to (or has) shed light on \exan{} assumptions.
These plans include any additional grounds for updating the report in case of sustained shifts in observations or metrics (\eg{} the effects of exogenous changes on system behaviors).
As such, this section must draw sustained reference to previous Reward Reports, including subsequent changes to the description, implementation, or evaluation, and what prompted these changes.
While previous sections outline how the system learns from data, this section tracks how organizations learn to oversee the system.
Its documentation is particularly important for defining \textit{accountability} for the system itself, those who manage it, and those responsible for completing periodic reports.
\begin{enumerate}[noitemsep,nolistsep]
\item \textbf{Reporting cadence}: The intended timeframe for revisiting the Reward Report. How was this decision reached and motivated?
\item \textbf{Update triggers}: Specific events (projected or historic) significant enough to warrant revisiting this report, beyond the cadence outlined above. Example triggers include a defined stakeholder group empowered to demand a system audit, or a specific metric (either of performance or oversight) that falls outside a defined threshold of critical safety.
\item \textbf{Changelog}: Descriptions of updates and lessons learned from observing and maintaining the deployed system.
This includes when the updates were made and what motivated them in light of previous reports. 
The changelog is the key difference between Reward Reports and other forms of machine learning documentation, as it successively reframes prior reports and reflects their intrinsically dynamic nature.
\end{enumerate}

\section{Examples}
\label{sec:examples}

Our aim with these examples is to illustrate the breadth and scope of questions that a Reward Report could engage with, and to demonstrate how Reward Reports can apply to both explicit and implicit RL systems in various design contexts.
Below, we focus on the Reward Report for the BlenderBot 3 chatbot deployed in August 2022~\cite{blenderbot2022}. 
We refer the reader to the appendix for the complete Reward Reports for BlenderBot 3 and several other examples including game-playing (DeepMind's MuZero~\cite{schrittwieser2020mastering}), content recommendation (MovieLens~\cite{harper2015movielens}), and traffic control (Project Flow~\cite{wu2021flow}).

\subsection{BlenderBot 3: A Chatbot Designed to Improve Over Time Through Feedback}
BlenderBot 3 is a recent chatbot "designed to improve its conversational skills and safety through feedback from people who chat with it, focusing on helpful feedback while avoiding learning from unhelpful or dangerous responses"~\cite{blenderbot2022}. 
To achieve this, the chatbot incorporates more than one type of feedback. 
First, it incorporates a significantly larger language model than its predecessorsat 175 billion parameters, unlocking new conversational capabilities via a larger capacity policy--a level of \textit{control feedback}. 
Second, it includes an interface for human users to provide real-time feedback on its conversational outputs if they are biased, inappropriate, or lack context--a kind of "reward signal" that uses behavior feedback for eventual model updates. 
Finally, its designers have articulated their high-level goal to release data about the chatbot's performance to the broader AI research community as a means to uncover new strategies for making future AI systems safer and more engaging to users--a kind of constructive form of exogenous feedback, beyond the technical specification scope. 

In parallel with BlenderBot 3's deployment, the designers made both its language model (OPT-175B) and associated model cards public to the AI research community. 
We have corresponded and worked with the designers to synthesize and interpret these resources as a basis for a Reward Report for BlenderBot 3.
This Reward Report includes the components outlined in Section 5, revealing potential interactions between different feedback types and associated questions pertaining to the system specification. 
These include:
\begin{itemize}
  \item What metrics (e.g. conversation length, ratio of negative vs. positive feedback) are being used to evaluate performance?
  \item How often will the language model be retrained?
  \item What outputs, if any, would compel designers to take BlenderBot 3 offline?
  \item Which stakeholders are responsible for the system's deployment, have a say in its specification, or have a veto over its public operation?
\end{itemize}
As a result, the BlenderBot 3 Reward Report does not merely aggregate model cards. 
It reveals that the documentation of feedback types requires a qualitative appraisal of system components, both in relation to each other and to the wider social context in which the system is intended to operate. 
This project entails a commitment to update the documentation over time as unintended types of feedback emerge and performance metrics are gradually refined.

\section{Discussion \& Conclusions}
\label{sec:discussion-and-conclusion}
The scale and complexity of contemporary optimization pipelines raise unique concerns not addressed by static reports and recent calls for documentation (e.g. those focusing on models or datasets).
ML deployments frequently consist of many moving parts and feedback channels that change over time, especially when the systems interact directly with multiple stakeholders like business customers and public users.
Reward Reports fill this gap, providing a framework for iterative deliberation over the time-evolution of a system and its feedback channels.
We have also demonstrated that the technical problem space and language of RL is useful for interpreting problems of fairness and safety in a reflexive and domain-specific manner.
While the utility of this framing will vary with application, Reward Reports will be of most use where a system is data-driven, and where actions have clear and automatic results.

Optimization-based policies in domains like school bus scheduling~\cite{schoolbus} or prison allocation~\cite{prison_allocation}, for example, are often not data-driven. 
However, it is not hard to imagine a future where these optimizations incorporate quantities predicted by statistical models. 
If this approach expands across domains, a standard mechanism for anticipating and deliberating over dynamic harms could become a critical component of governance. 

The pace of academic research also suggests that in the near future, implicit or explicit RL systems will be more effective, deployed in more impactful user-facing applications, and fine-tuned in `real-time' rather than re-trained daily or weekly.
The complexities of real-time training are compounded with the addition of human-in-the-loop data collection, such as in reinforcement learning from human feedback~\cite{kirk2023personalisation}. 
The resulting feedback loops will bring more and more problems into the purview of the RL lens, further motivating the need for Reward Reports.

Reward Reports could also be of use for human-in-the-loop ML deployments where actions do not have automatic impact.
Clinical decision support systems can be data-driven, but the clinician ultimately determines how system recommendations are implemented.
In this case, the human computer interaction (HCI) component of such a system could distort
the interface between recommendations and human judgment in ways that are not captured within a pure RL lens.
For example, a doctor is more likely to defer to an incorrect recommendation by an algorithm when it is accompanied by a paragraph of reasoning than without~\cite{jacobs2021recommendations}. However, the deliberation over system feedback made possible by Reward Reports could still elucidate unforeseen harms.

These examples all point to a deeper truth: designing systems to promote societal good is an increasingly dynamic problem, and it needs to be deliberated about as such.
Reward Reports enact forms of documentation commensurate with the feedback-laden systems whose dynamics --not just models or data-- are a critical object of concern.

\begin{acks}
The authors would like to acknowledge the Center for Human Compatible AI, the Center for Long Term Cybersecurity for their support, and the Mozilla Foundation.
The authors also thank Soham Mehta for his contributions to writing BlenderBot3 Reward Report and to the BlenderBot3 team at Meta AI for aiding us in this project.
\end{acks}

\bibliographystyle{ACM-Reference-Format}
\bibliography{main}

\clearpage

\appendix

\onecolumn
\section{Empty Reward Report Template}

We also include an empty template for a Reward Report, including descriptions of the content for each section.
\\

\begin{center}
    \includegraphics[
        scale=0.60,
        page=1,
        frame,
    ]{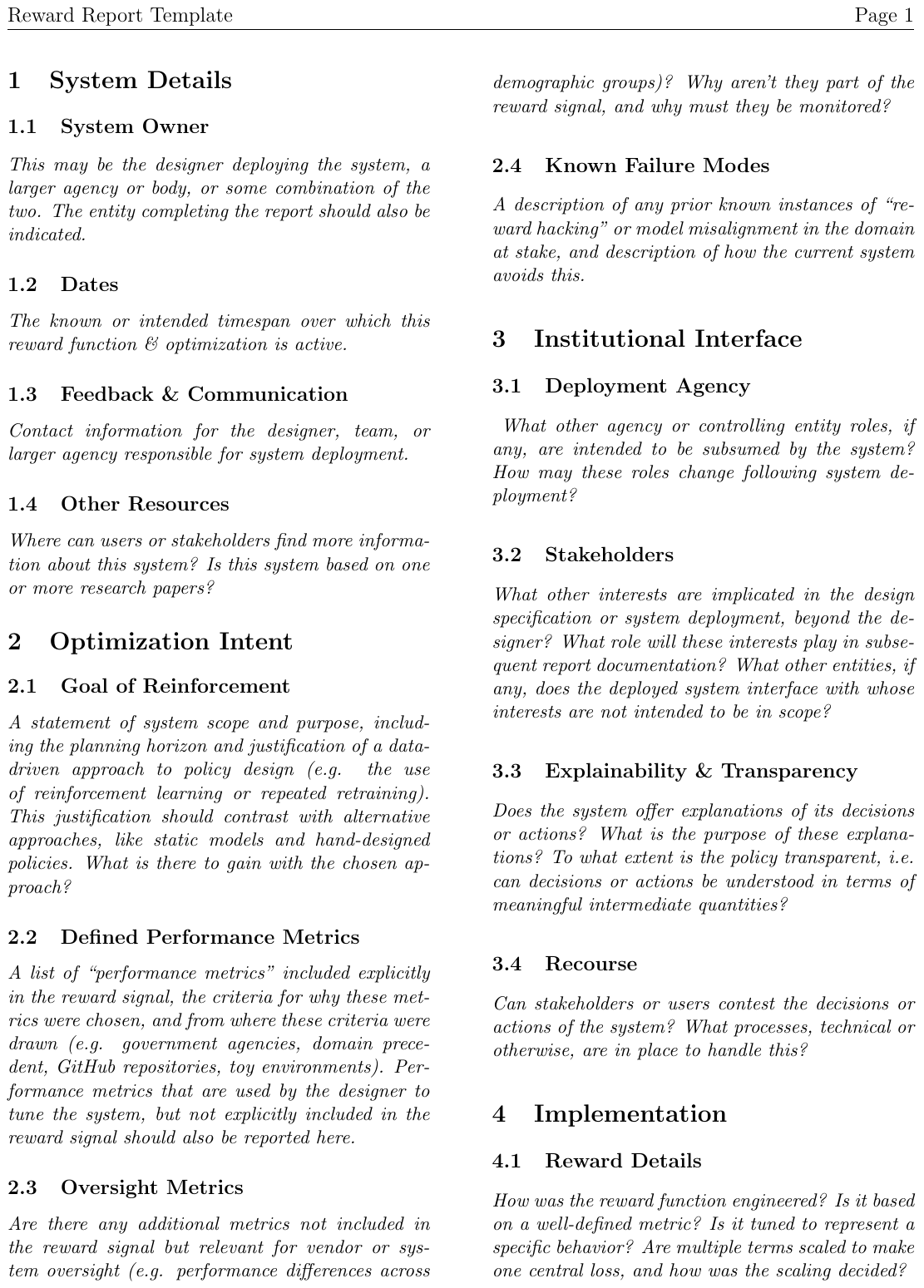}
\end{center}

\clearpage

\includepdf[
    pages=2-last,
    scale=0.85,
    frame=true,
    delta=0.5cm 0.5cm,
]{examples/appendix-reward-report-template2.pdf}

\section{Example Reward Report - BlenderBot}

\begin{center}
    \includegraphics[
        scale=0.63,
        page=1,
        frame,
    ]{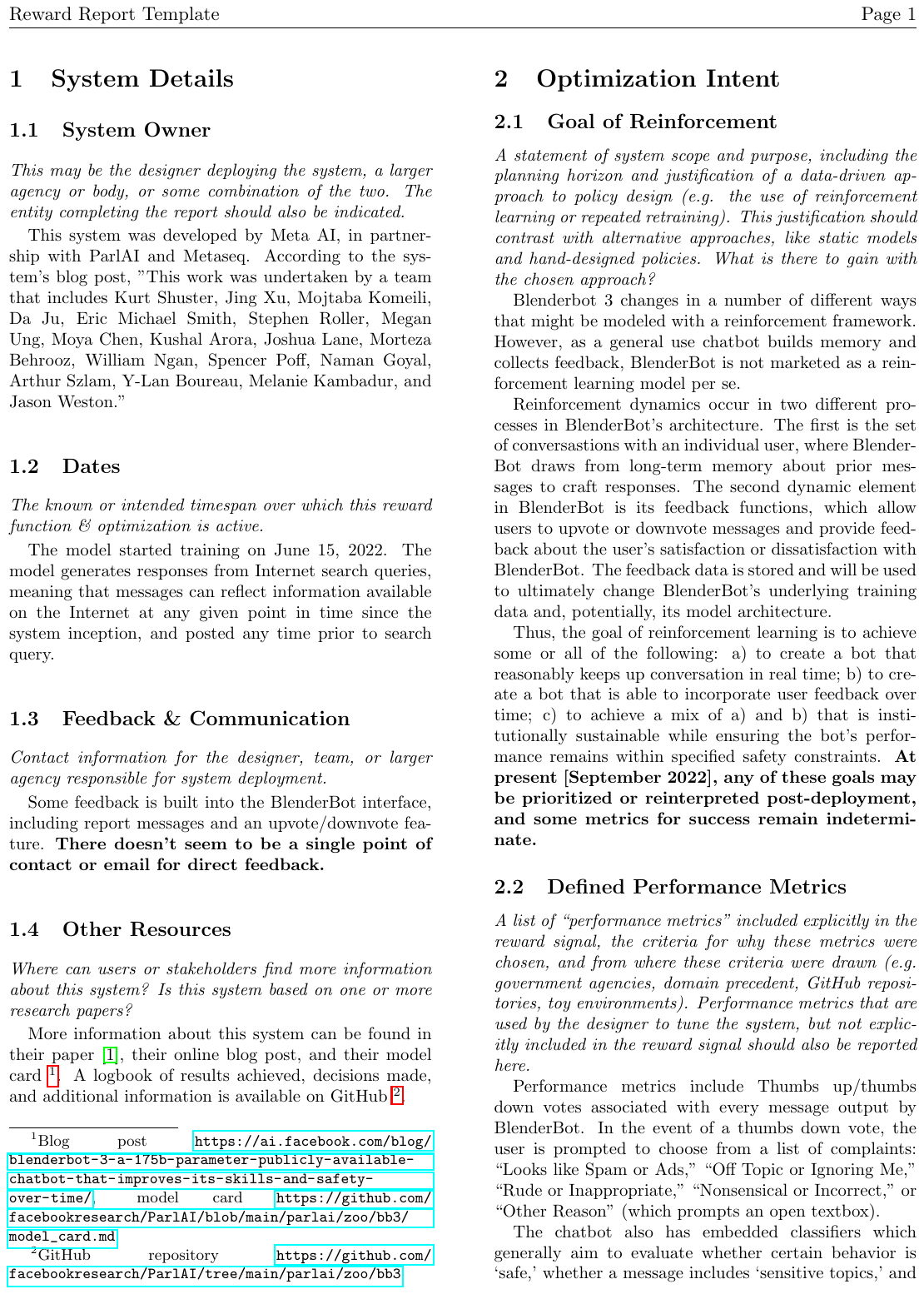}
\end{center}

\clearpage

\includepdf[
    pages=2-last,
    scale=0.85,
    frame=true,
    delta=0.5cm 0.5cm,
]{examples/BlenderBot_Reward_Report_Feb23.pdf}

\section{Additional Examples}

\subsection{Example Reward Report - Project Flow}
\label{app:projectflow}

Project Flow is an autonomous vehicle testbed that allows using deep reinforcement learning to control and optimize traffic across in roadway networks \cite{wu2021flow}. 
Inspired by recent work with using Project Flow~\cite{Kreidieh2018Dissipating}, we sketch a hypothetical deployment of an RL policy designed for dissipating stop-and-go traffic waves at a freeway exit, including several iterations of the Reward Report documented in the accompanying changelog.
The changelog shows various problems that arise with the resulting problem dynamics, including an expansion of the planning horizon, the addition of new oversight metrics, stakeholder complaints, and requisite institutional shifts to cope with changes to the specification and application domain.

\smallskip

\begin{center}
    \includegraphics[
        scale=0.58,
        page=1,
        frame,
    ]{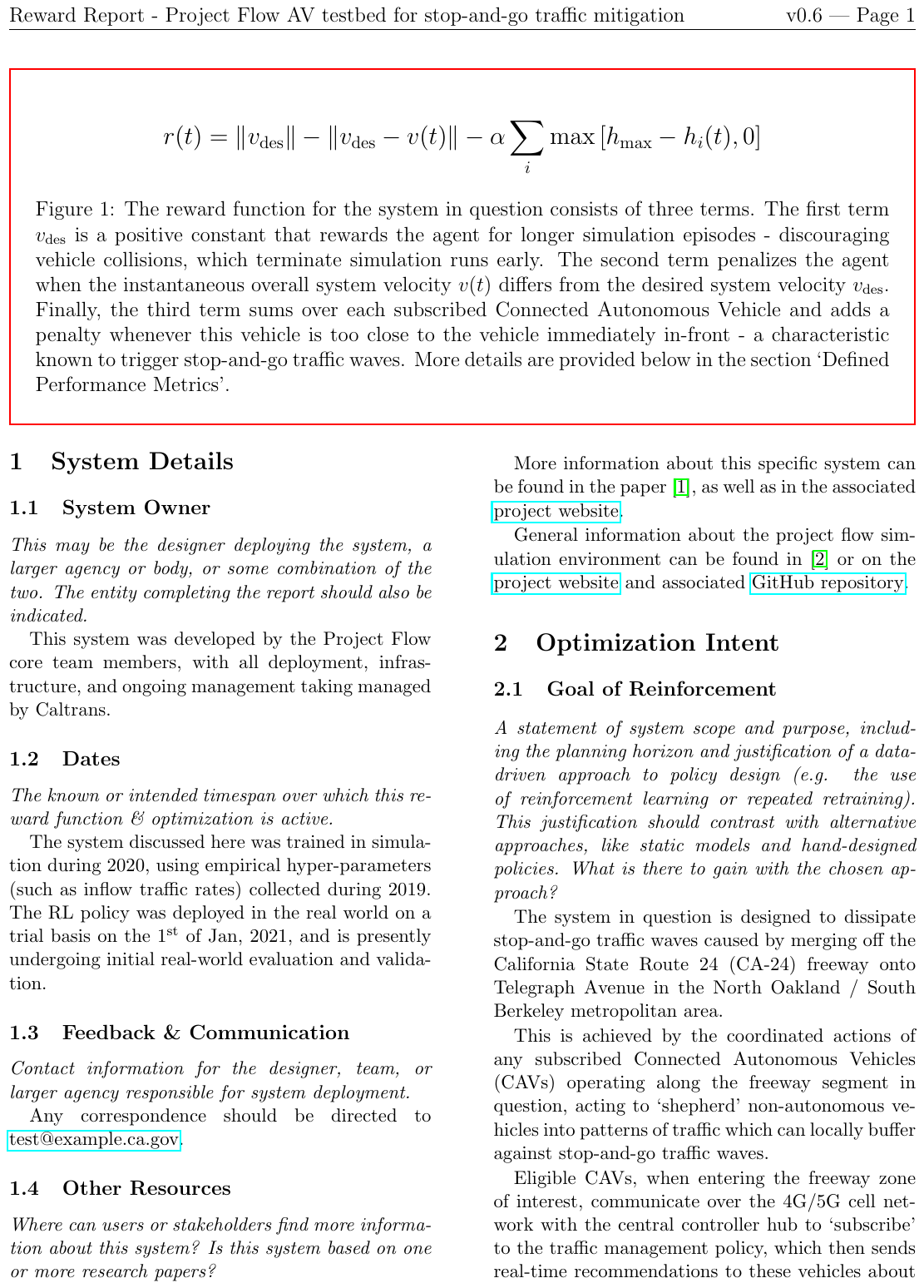}
\end{center}

\clearpage

\includepdf[
    pages=2-last,
    scale=0.85,
    frame=true,
    delta=0.5cm 0.5cm,
]{examples/appendix-reward-report-projectflow2.pdf}

\subsection{Example Reward Report - MovieLens}
\label{app:movielens}

The purpose of MovieLens is to match users to personalized movie recommendations based on ratings of other movies previously entered by the user \cite{harper2015movielens}.
Unlike the other example systems we discuss, MovieLens is a static preference model generated through supervised learning.
However, because of the system's age (initial release in 1997) and its repeated retraining, it can be interpreted as an RL system that is learning a ranking policy that must adapt to a changing environment.
The changelog documents the actual historical updates to the model prompted by changes to the environment, including new interfaces, user-base size, optimization parameters, user-generated content, and major dataset publications.
This example Reward Report is based on the history of the MovieLens project published in \cite{harper2015movielens}.

\smallskip

\begin{center}
    \includegraphics[
        scale=0.63,
        page=1,
        frame,
    ]{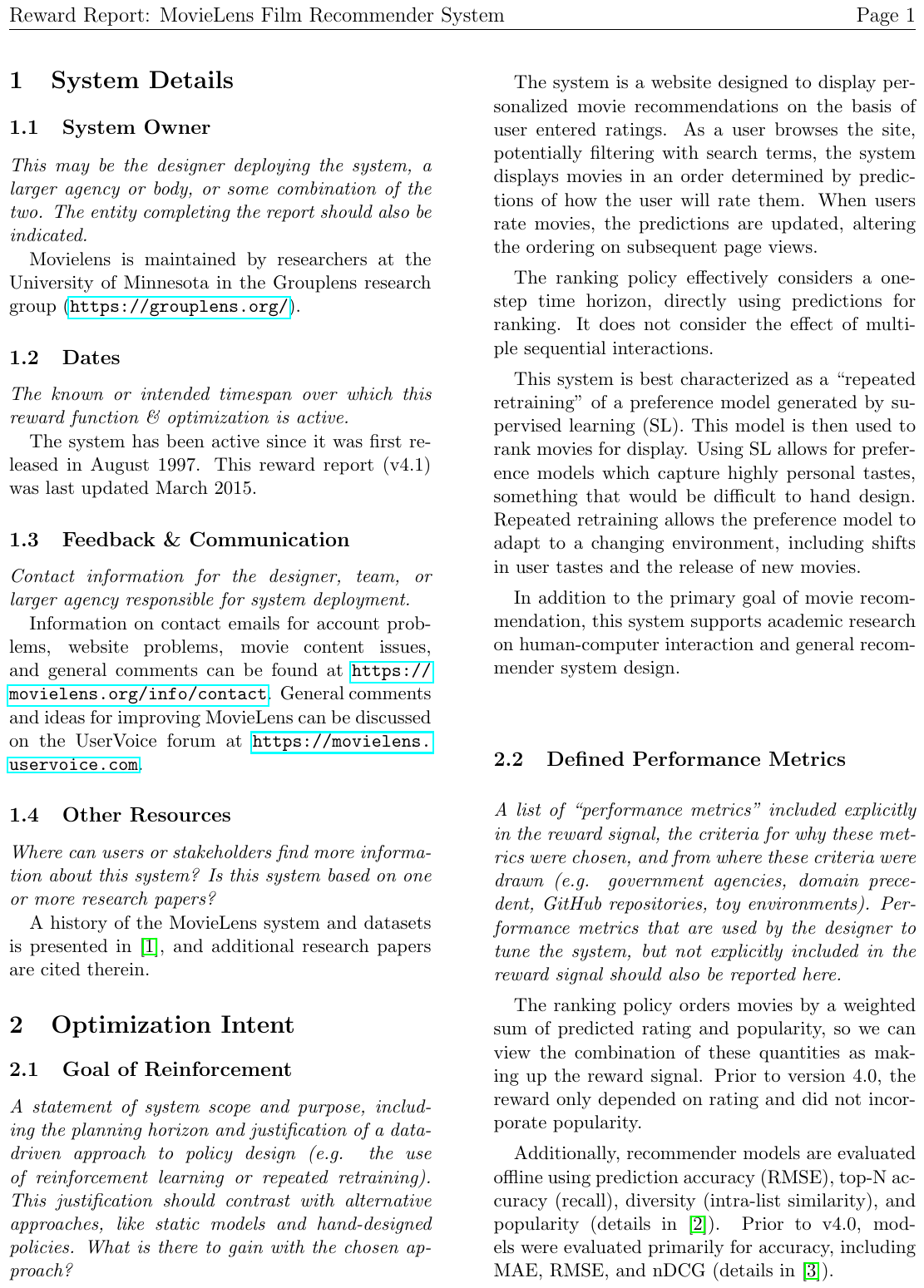}
\end{center}

\clearpage

\includepdf[
    pages=2-last,
    scale=0.85,
    frame=true,
    delta=0.5cm 0.5cm,
]{examples/appendix-reward-report-movielens2.pdf}

\subsection{Example Reward Report - MuZero}
\label{app:muzero}

The purpose of MuZero (and its preceding systems, AlphaGo and AlphaZero) is to improve state-of-the-art performance in the games of chess, Go, shogi, and a benchmark suite of Atari games \cite{silver2016mastering}.
We provide a Reward Report that documents the evolution of the system through these successive stages of development, including changes in the design motivation and performance metrics, as well as more extensive use of reinforcement learning.
\smallskip

\begin{center}
    \includegraphics[
        scale=0.63,
        page=1,
        frame,
    ]{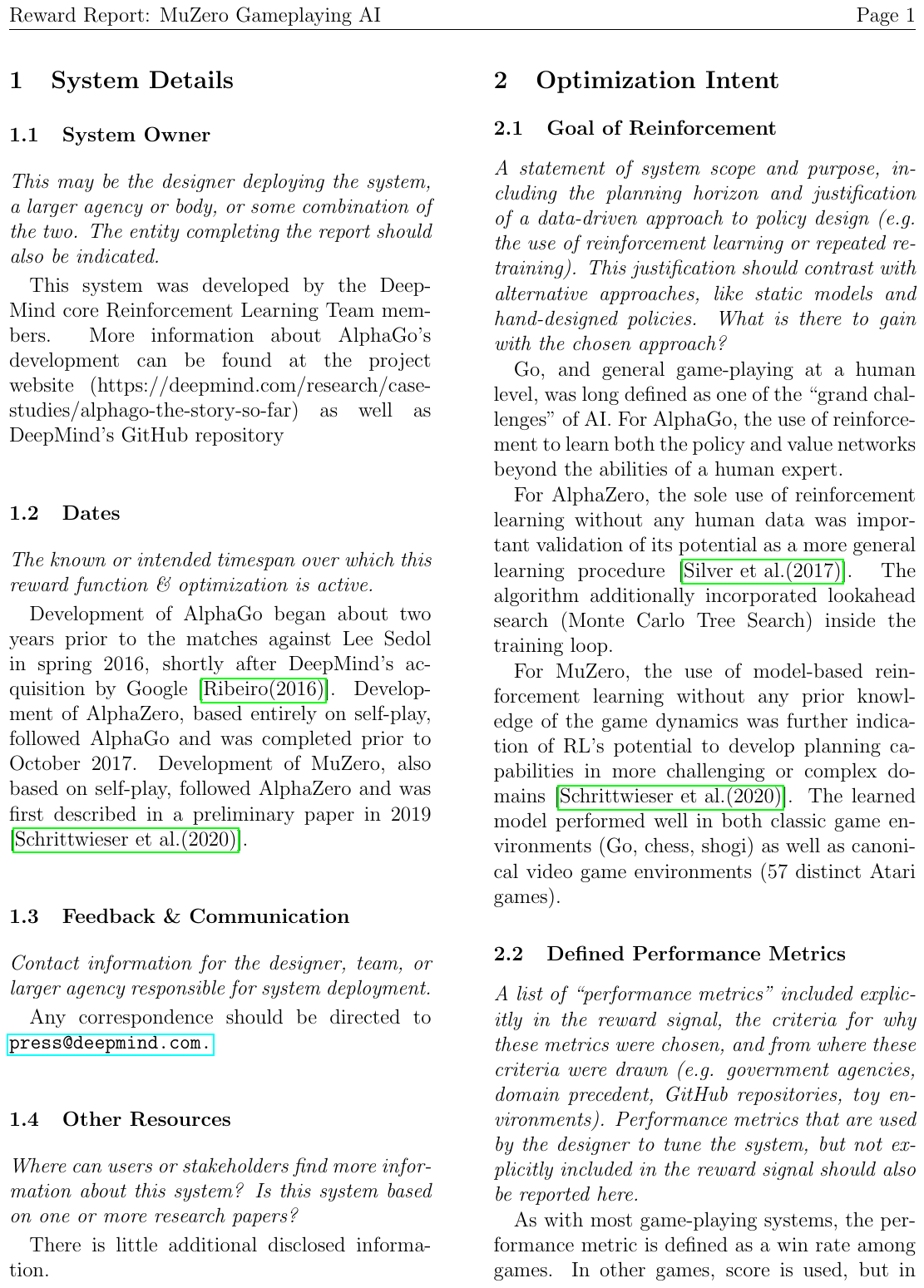}
\end{center}

\clearpage

\includepdf[
    pages=2-last,
    scale=0.85,
    frame=true,
    delta=0.5cm 0.5cm,
]{examples/appendix-reward-report-muzero2.pdf}

\end{document}


\begin{figure*}[b]
\begin{center}

    \includegraphics[height=7cm]{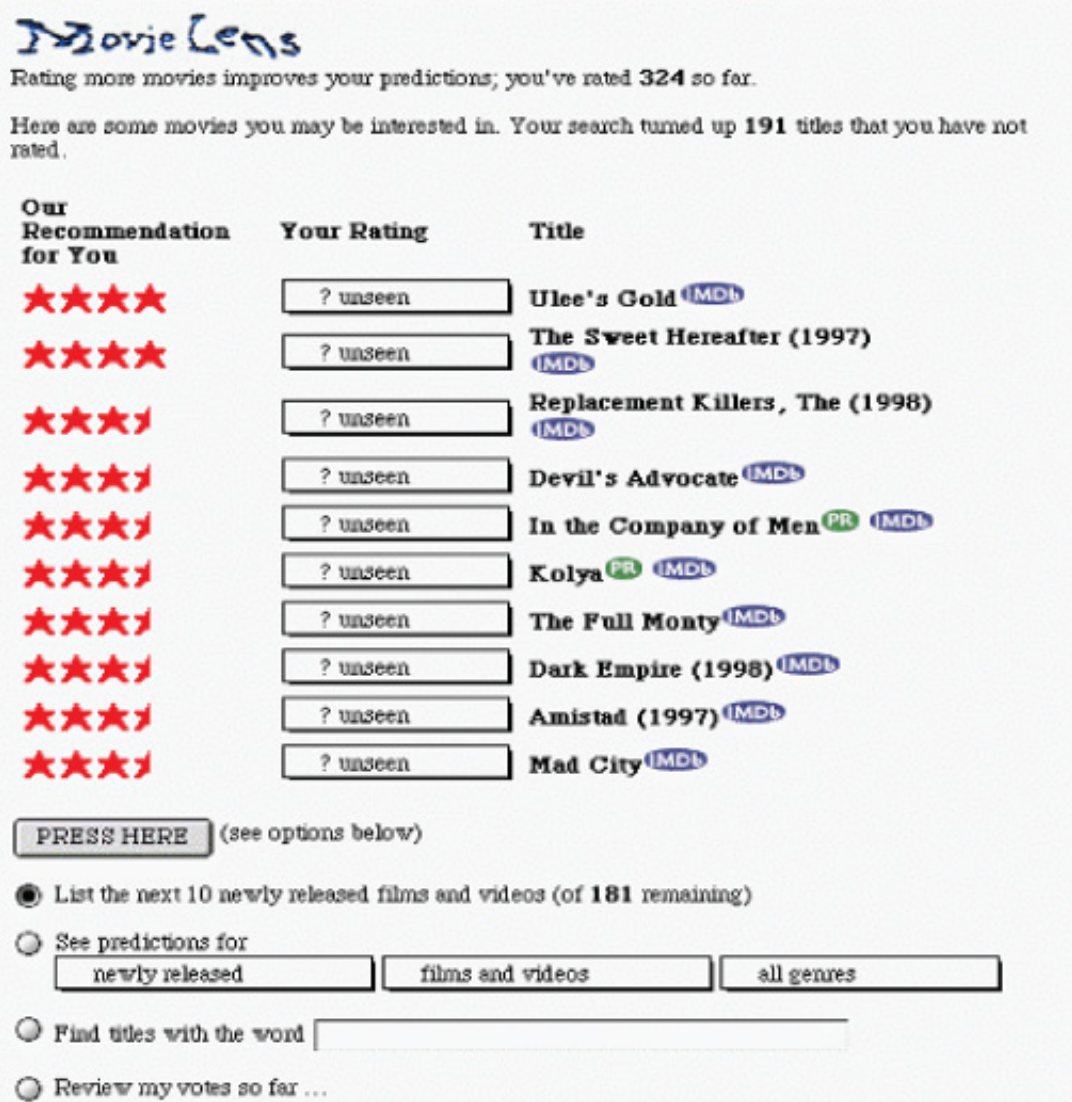}~
    \includegraphics[height=7cm]{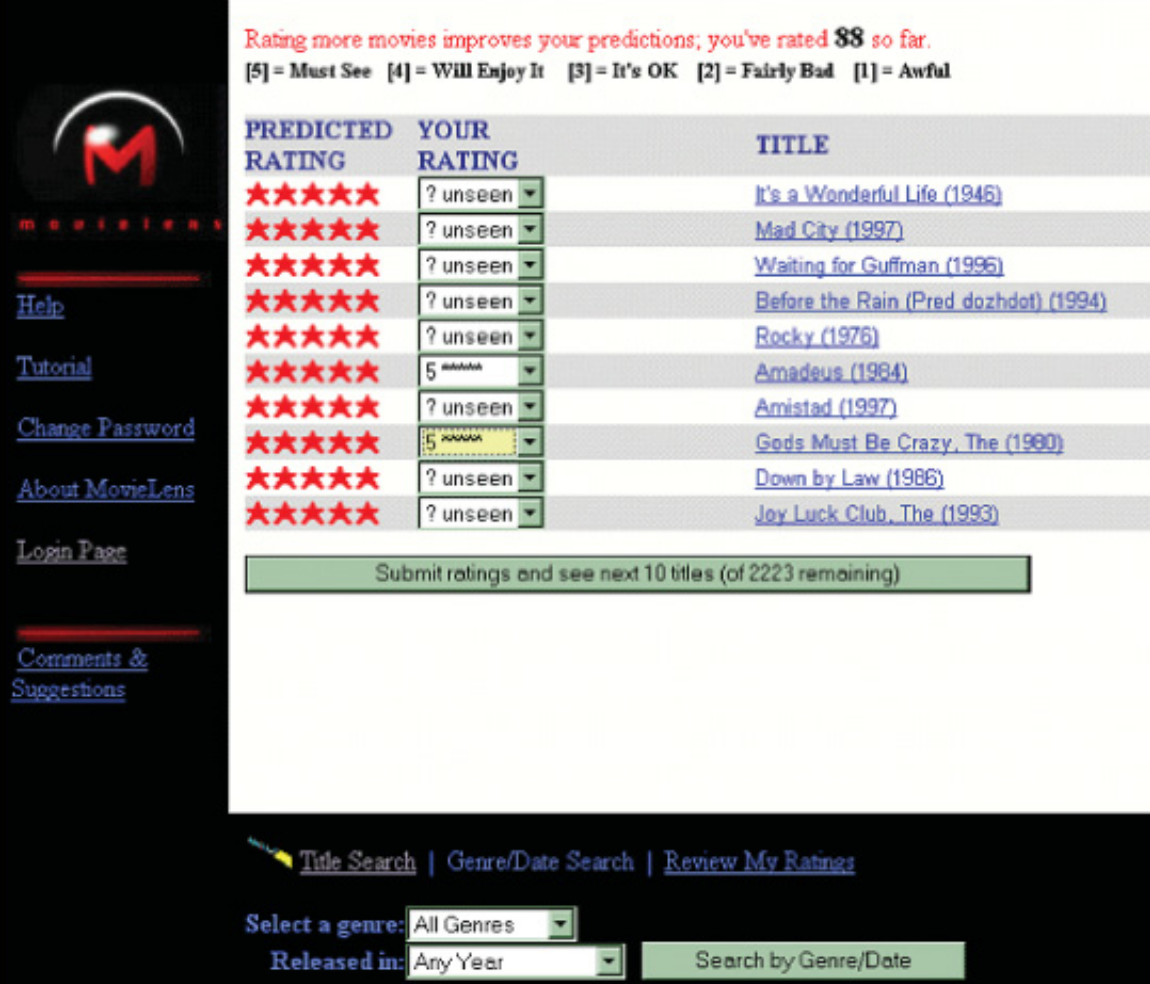}\\
    \includegraphics[height=7cm]{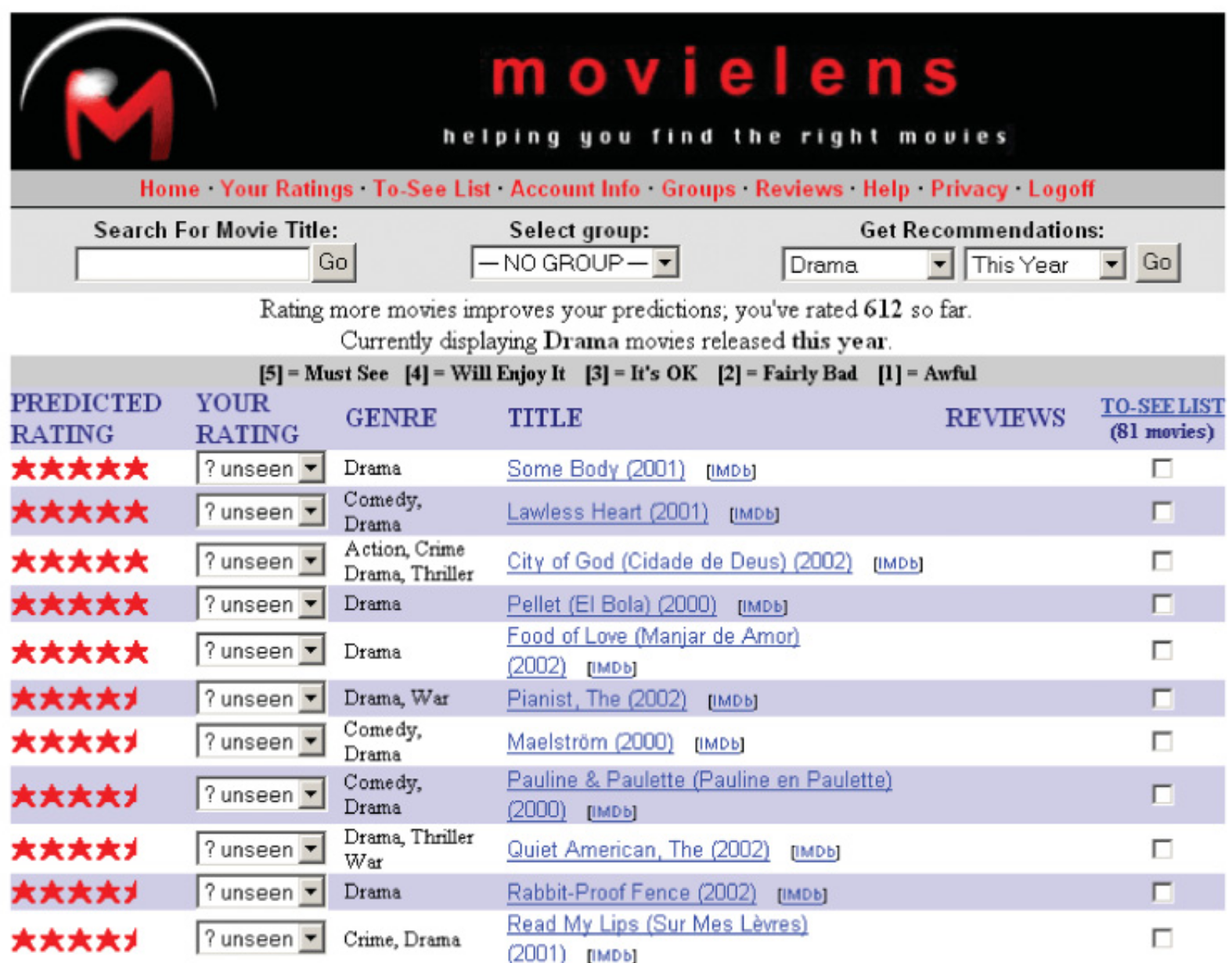}~
    \includegraphics[height=7cm]{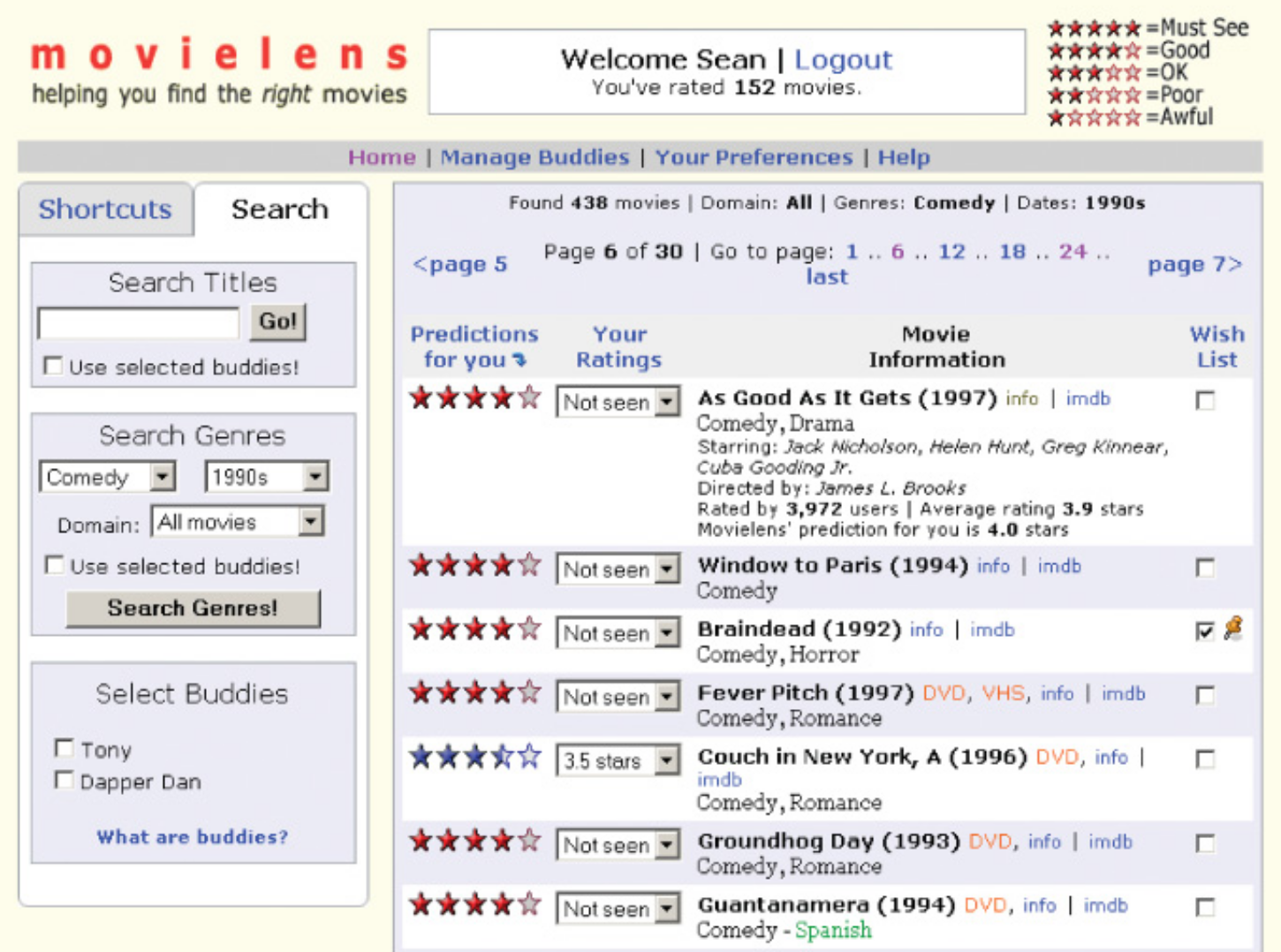}
     \includegraphics[height=7cm]{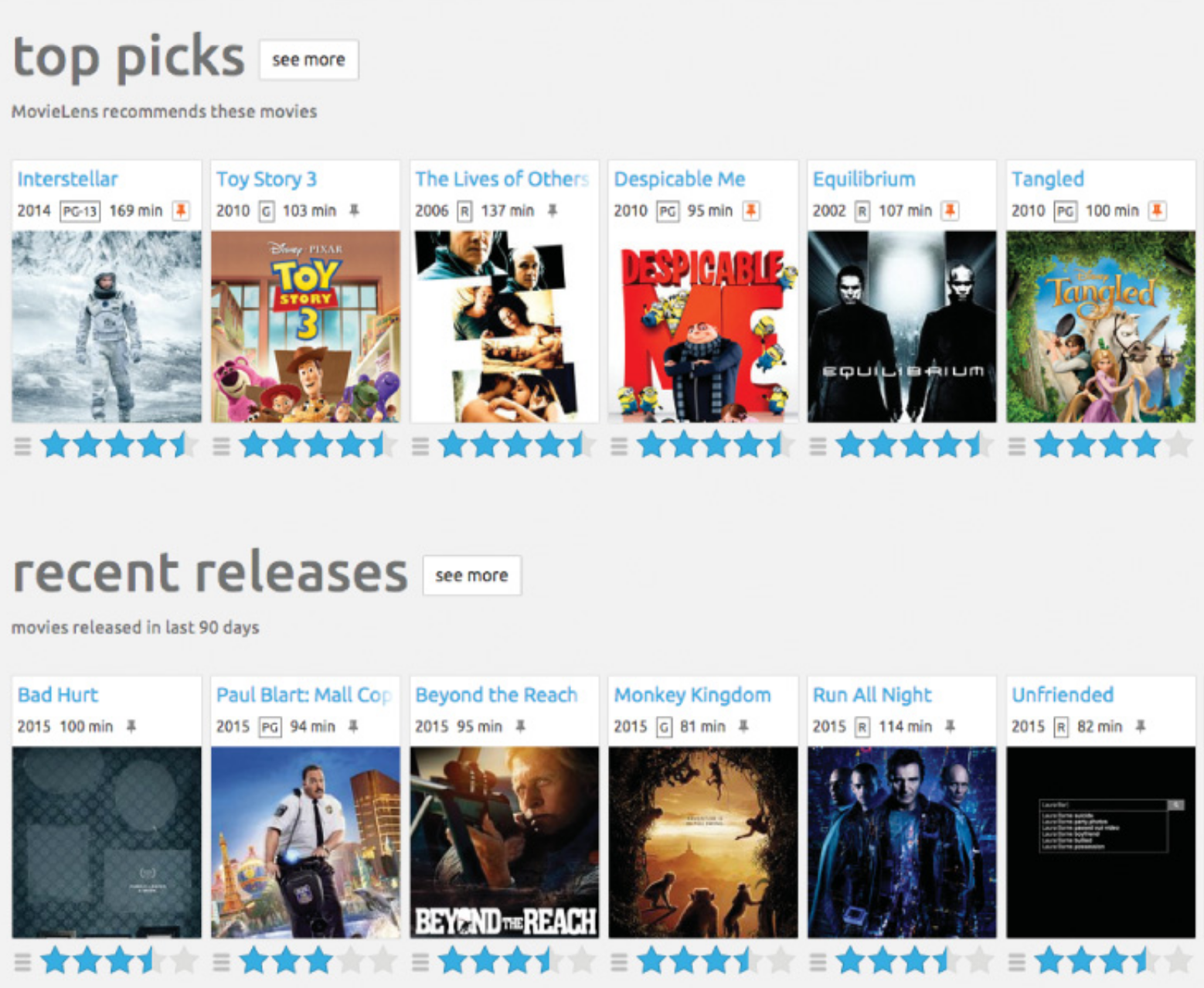}
        
\end{center}
    \caption{%
        The MovieLens recommender system interface v0-v4.
    }\label{fig:interface}
\end{figure*}

\SystemDetails{
Movielens is maintained by researchers at the University of Minnesota in the Grouplens research group (\url{https://grouplens.org/}).
}{
The system has been active since it was first released in August 1997.
This reward report (v4.1) was last updated March 2015.
}{
Information on contact emails for account problems, website problems, movie content issues, and general comments can be found at
\url{https://movielens.org/info/contact}. General comments and ideas for improving MovieLens can be discussed on the UserVoice forum at \url{https://movielens.uservoice.com}.
}{
A history of the MovieLens system and datasets is presented in~\cite{harper2015movielens}, and additional research papers are cited therein.
}

\OptIntent{
The system is a website designed to display personalized movie recommendations on the basis of user entered ratings.
As a user browses the site, potentially filtering with search terms, 
the system displays movies in an order determined by predictions of how the user will rate them.
When users rate movies, the predictions are updated, altering the ordering on subsequent page views.

The ranking policy effectively considers a one-step time horizon, directly using predictions for ranking. It does not consider the effect of multiple sequential interactions.

This system is best characterized as a ``repeated retraining'' of a preference model generated by supervised learning (SL).
This model is then used to rank movies for display.
Using SL allows for preference models which capture highly personal tastes, something that would be difficult to hand design.
Repeated retraining allows the preference model to adapt to a changing environment, including shifts in user tastes and the release of new movies.

In addition to the primary goal of movie recommendation, this system supports academic research on human-computer interaction and general recommender system design.
}{
The ranking policy orders movies by a weighted sum of predicted rating and popularity, so
we can view the combination of these quantities as making up the reward signal.
Prior to version 4.0, the reward only depended on rating and did not incorporate popularity.

Additionally, recommender models are evaluated offline using prediction accuracy (RMSE), top-N accuracy (recall), diversity (intra-list similarity), and popularity (details in~\cite{ekstrand2015letting}).
Prior to v4.0, models were evaluated primarily for accuracy, including MAE, RMSE, and nDCG (details in~\cite{ekstrand2011rethinking}).
}{
Metrics which are monitored but not incorporated into the policy or model include the number of users, number of movies, number of entered ratings, monthly active users, and the number of logins for each user.
These indicators of overall system operation are not targets for optimization.

}{

No instances of reward hacking or misalignment have been observed. 
Because the system allows for explicit user input (search terms, model selection), errors in rating predictions do not prevent users from finding and rating movies.

}

\Interfaces{
MovieLens was released due to the shuttering of EachMovie in 1997, a movie recommendation site hosted by DEC.
It was developed and is maintained by Grouplens, a research group at University of Minnesota.
}{
One interface of interest is the technology that powers the recommendation engine. Currently, it is powered by Lenskit, an open source framework developed to promote reproducability and openness in the recommendation systems community~\cite{ekstrand2011rethinking}. 
Previously in v3.0-v3.4, the recommendations were powered by MultiLens, another open source recommendation engine. 
MultiLens replaced Net Perceptions (v1.1-v2.0), a recommendations systems company cofounded in 1996 by GroupLens faculty and students and sold in 2004~\cite{usatoday}.
The recommendation model in v0.0-v1.0 was originally developed by GroupLens for personalized Usenet news recommendation~\cite{konstan1997grouplens}.

Another relevant interface is with The Movie Database, a free and open source user editable movie database for plot summaries, movie artwork, and trailers.
Previously, from in v3.4-v4.0, MovieLens integrated with the Netflix API to display movie posters and plot synopsis on the movie details page. However, Netflix eventually discontinued its API support.

An important stakeholder is the Movielens users. 
Soliciting user judgements and opinions is often a key element in determining if an experimental
change is successful.
Additionally, one-off user studies (with participants recruited from email) are used to test features that are not ready to scale or integrate into the main user interface.

Finally, a key stakeholder is the researchers: both in Grouplens and the in the community more broadly. 
The openness of users to experiments on a broad range of features has enabled GroupLens research in many different areas on the Movielens platform.
The regular release of anonymized datasets of movie ratings is important to the broader machine learning, data science, and information retrieval communities.

A potentially relevant group of stakeholders is movie producers. 
However, because Movielens is relatively small and isolated from larger commercial endeavors, it has limited impact on movie studios and production, so their interests are not in scope.
}{
The system displays predicted ratings alongside movies, explaining the movies position within a list, and suggesting to the user whether or not they will like the movie. 
The ranking policy is easily understood as a weighted combination of predicted rating and popularity.
However, the computation of predicted ratings is more complex. 
Some available models are more easily explained to users than others (e.g. nearest neighbors vs. matrix factorization).
However, the details are well documented in publicly available research papers~\cite{ekstrand2015letting}, and researchers respond to user requests for explanation on the UserVoice discussion board~\cite{explainml}.
}{
By entering ratings, users are able to affect their preference models to hopefully become more accurate.
Additionally, the movies displayed by the system are sourced from The Movie Database, which is user-editable. 
(Previously in v3.2-v3.5, users could add and edit movies to MovieLens directly.)
Furthermore, the current version of the system allows users to
choose between three recommender models.
Finally, users can make suggestions and requests directly to designers on the UserVoice forum.
}

\Implementation{
The reward is a weighted sum: 
$$0.9\cdot \mathrm{rank}(\widehat r_{ui}) + 0.1\cdot \mathrm{rank}(p_i)$$
where $\widehat r_{ui}$ is the predicted rating of movie $i$ by user $u$, $p_i$ is the number of ratings movie $i$ has recieved in the past 10 days, and $\mathrm{rank}$ normalizes input, returning $1$ for the largest (across all movies) and $0$ for the smallest. 
This blending is the result of empirical evidence that it improves user satisfaction.

}{
The system handles approximately 250k users and 30k movies. 
These numbers have grown over the years.
In 1999 (v1.1), MovieLens received attention from the mass media, causing an increase in user signups. 
Since then, the user growth has been stable (20-30 signups per day), largely the result of word-of-mouth or unsolicited press.
Early on, the movie database was hand-curated and primarily contained movies with wide theatrical release in the United States.
In v3.2-v3.5, MovieLens added the ability for users to edit and add movies.
Since v4.0, MovieLens uses The Movie Database, a free and open source user editable movie database.

The actions taken by the system are page displays of 10 movies in a ordered list, where pages can be perused by arrows.
The views can be explicitly filtered with search terms like year and genre; these explicit inputs this make up a component of the observation.
The second component is the entered ratings in the form \texttt{<user\_id, movie\_id, rating, timestamp>}.

There are three potential sources of dynamics in this environment: the addition of new movies, the joining and departing of users, and the preferences that users have for movies.
Because this system effectively uses a planning horizon of 1, none of these dynamics are explicitly accounted for.
This is appropriate, as the goal of MovieLens is not to shift broad patterns of movie consumption. Though the movies, users, and preferences may change over time, these changes are more likely to be due to external factors than feedback with the MovieLens system.
Additionally, the data collected by MovieLens is not fine-grained enough to detect such impacts of feedback.
}{

Ratings are entered by users via clicks on a star graphic, and can take values 0.5-5 in half integer increments.
Prior to v3.0, ratings took values in integer increments. The increased granularity was the most requested feature in a user survey.
Prior to v4.0, ratings were entered through a drop-down menu, and the meaning of rating values was described in a legend at the top of the page (see Figure~\ref{fig:interface}).

A possible source of bias in the measured ratings is due to anchoring effects, due either to the displayed predicted rating or due to the historically provided movie rating legend.
However, broad trends in rating values did not change when the legend was removed in v4.0

Finally, the recorded timestamp represents when a user adds a particular rating rather than when they watched a movie. 
This limits the ability of the system to detect the impacts of its own recommendations.

}{
The policy selects a page view to present to the user based on explicitly provided input and rating data.
First, explicit input is used to filter the list of movies. 
Then, the recommender model is used to predict a user's ratings of these movies.
Finally, the movies are displayed in order of these predicted ratings, blended with a popularity factor.

The main component of the policy is therefore the recommender model. This model is user-selectable, so that users can choose between a non-personalized baseline, a preference elicitation model intended for new users, an item-item collaborative filtering model, or a matrix factorization model. Further details on how these models are trained is available in~\cite{ekstrand2015letting}.
Previously in v3.0-3.5, the recommender was fixed as an item-item collaborative filtering model.
Prior to that in v1.0-2.0, the model was a user-user collaborative filtering model.
}{
All user rating data is stored by MovieLens and used by the recommender models to make rating predictions. 
When a user enters a new rating, it immediately impacts their rating predictions, since the ``input'' to the recommender changes. Less frequently, the ratings are used to update the parameters of the recommender models.
An anonymized subset of this data is also periodically released for use by the wider research community.

The dataset of user ratings is likely biased.
There is sampling bias due to the fact that users only rate movies that 1) appear on a page and 2) that they have watched. These factors are directly and indirectly  impacted by the MovieLens system itself.
The fact that users can explicitly filter pageviews with search terms mitigates these effects, but it is unlikely that it removes them.

The initial Movielens system was trained on a public dataset from EachMovie of approximately 2.8 million ratings from 72k users across 1.6k movies, but this has since been discarded. The dataset was retired by HP in October 2004, and due to privacy concerns, it is no longer available for download.

}{
The most prevalent limitation of this system is that it does not plan over a long horizon and therefore does not consider the effects of dynamics.
While a more complex policy would allow the system to adapt to ordering effects,
the resulting temporal dependence would complicate the ability to users to reliably navigate the movie database.
Furthermore, users do not always enter movie ratings immediately after watching a movie, instead sometimes entering batches of ratings for movies that they watched in the past.
}{
The system cannot provide reliable recommendations until users provide a minimum number of ratings. 
This problem is avoided by the interface design: when a user joins the site, they express their preferences over several displayed clusters of movies. These preferences are used, in combination with the rating profiles of other users, to generate a psuedo-rating profile for the new user. Further description is available in \cite{chang2015using}.

This preference elicitation process replaced a minimum movie requirement.
Previously, until a user rated a minimum number of movies, the front page would display 10 movies at a time.
From v0-v3, the minimum number was 5, and of the 10 movies per page,
nine were randomly selected from the database and one from a hand-designed list of recognizable titles.
In v3, the minimum number was 15, and the 10 movies were selected for their popularity, excluding the top 50-150 movies. This increased requirement was due to the needs of an item-item (rather than user-user) collaborative filtering algorithm.
The switch to a preference elicitation process was motivated by the observation that the 15 rating requirement was too arduous, taking users an average of 6.8 minutes to complete and 12.6\% of users failing to complete it.

}

\begin{figure*}[t]
\begin{center}

    \includegraphics[width=\textwidth]{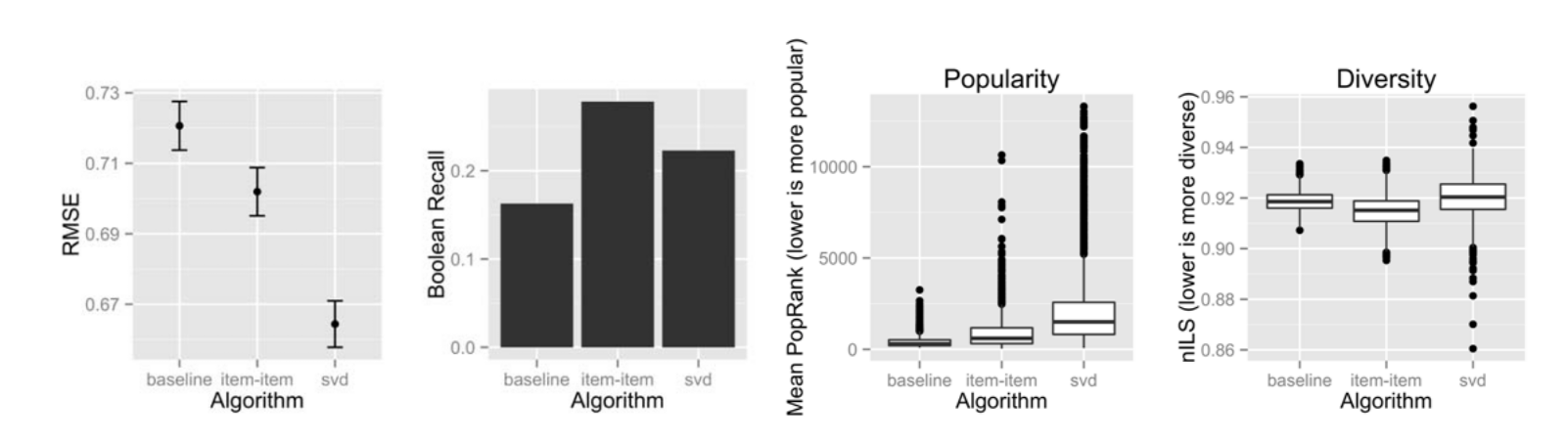}
        
\end{center}
    \caption{%
        Offline evaluation of recommender models from~\cite{ekstrand2015letting}.
    }\label{fig:eval}
\end{figure*}

\Evaluation{
The primary evaluation is to consider various properties of recommender models on offline datasets. This includes many of the publicly released MovieLens datasets, which are described in detail in~\cite{harper2015movielens}.
}{
This offline evaluation includes prediction accuracy (RMSE), top-N accuracy (recall), diversity (intra-list similarity), and popularity.
Detailed evaluations are available in~\cite{ekstrand2015letting}, and key quantities are displayed in (Figure~\ref{fig:eval}).
}{

Offline evaluation metrics (like top-N accuracy) were chosen to align with the ranking setting. 
While the offline evaluations are imperfect (due to dataset biases), the system appears to work well ad no unexpected behaviors have been observed.
}{
N/A
}

\Maintenance{
This report is updated whenever there is a major system update, either to the user interface or the backend.
Such updates will occur periodically, coinciding with research initiatives.
}{
If a large change is observed in oversight metrics, or if many users express dissatisfaction on the UserVoice forum, the system design will be revisited by the researchers who maintain it. If an update is deemed necessary, this report will be updated.

}{
The versions of this report are enumerated as vX.Y where X corresponds to the user interface version and Y corresponds to major changes within interfaces.

\begin{itemize}
    \item v0.0 (August 1997) Initial release.
    \item v0.1 (April 1998) The ML 100K dataset is released, covering 9/1997–4/1998.
    \item v1.0 (September 1999) Update to v1 interface.
    \item v1.1 (November 1999) Media exposure causes an increased number of users. Switch from GroupLens to Net Perceptions recommender model.
    \item v2.0 (February 2000) Update to v2 interface. Additional movie metadata and reviews added to movie details pages.
    \item v3.0 (February 2003) Update to v3 interface. Switch from Net Perceptions user-user recommender to MultiLens item-item recommender. Ratings now in half-star (rather than full) increments. Require that users rate at least 15 movies before receiving recommendations. The ML 1M dataset is released, covering  4/2000–2/2003.
    \item v3.1 (June 2005) Added discussion forums to site.
    \item v3.2 (September 2008) Added feature so that users can add movies to database.
    \item v3.3 (January 2009) The ML 10M dataset is released, covering  1/1995–1/2009.
    \item v3.4 (Spring 2009) Netflix API integration for poster art and synopsis.
    \item v3.5 (January 2012) Switch from Multilens to Lenskit recommender (still item-item).
    \item v4.0 (November 2014) Update to v4 interface. Rating interface combined with ``predicted rating'' star graphic to accept click events. Switch to user-selectable recommender model. Legend describing the meanings of ratings and dropdown menu removed. Drop minimum rating requirement in favor of group-based preference elicitation. Integration with The Movie Database for plot summaries, movie artwork, and trailers.
    \item v4.1 (March 2015) The ML 20M dataset is released, covering  1/1995–3/2015.
    Moving forward, MovieLens will make public additional nonarchival datasets: \texttt{latest} which is unabridged for completeness and \texttt{latest-small} for educational use.
\end{itemize}

}

\bibliographystyle{IEEEtran}
\bibliography{main}


\SystemDetails
{This system was developed by the DeepMind
core Reinforcement Learning Team members.
More information about AlphaGo's development can be found at the project website (https://deepmind.com/research/case-studies/alphago-the-story-so-far) as well as DeepMind's GitHub repository}
{Development of AlphaGo began about two years prior to the matches against Lee Sedol in spring 2016, shortly after DeepMind's acquisition by Google \cite{pcworld}.
Development of AlphaZero, based entirely on self-play, followed AlphaGo and was completed prior to October 2017.
Development of MuZero, also based on self-play, followed AlphaZero and was first described in a preliminary paper in 2019 \cite{schrittwieser2020mastering}. }
{Any correspondence should be directed to
\url{press@deepmind.com.}}
{There is little additional disclosed information.}

\OptIntent
{Go, and general game-playing at a human level, was long defined as one of the ``grand challenges'' of AI. 
For AlphaGo, the use of reinforcement to learn both the policy and value networks beyond the abilities of a human expert. 

For AlphaZero, the sole use of reinforcement learning without any human data was important validation of its potential as a more general learning procedure \cite{silver2017mastering}.
The algorithm additionally incorporated lookahead search (Monte Carlo Tree Search) inside the training loop.

For MuZero, the use of model-based reinforcement learning without any prior knowledge of the game dynamics was further indication of RL's potential to develop planning capabilities in more challenging or complex domains \cite{schrittwieser2020mastering}.
The learned model performed well in both classic game environments (Go, chess, shogi) as well as canonical video game environments (57 distinct Atari games).}
{As with most game-playing systems, the performance metric is defined as a win rate among games.
In other games, score is used, but in one-versus-one games win rate is the only direct metric.
To better capture the uncertainty of playing varying opponents, this win rate is translated into a running Elo rating system.}
{Some other performance metrics are not included in the specification, but are monitored for the purpose of evaluating system effects on the domain:

\begin{itemize}
    \item Absolute opponents' world rankings - following their public games, versions of AlphaGo and AlphaZero were considered to possibly improve the skill levels of expert human opponents, as measured by those players' absolute world ranking.
    If humans played better after playing AlphaGo, this was to be seen as a positive effect of the system's influence on the game of Go.
    Fan Hui, following his games against AlphaGo, claimed it made him a better played and accredits his world ranking jump from 600 to 300 in three months to training against it \cite{fan}.
    \item Qualitative changes in playstyle - following their public games, versions of AlphaGo were considered to possibly  influence the playstyle of expert human opponents, as interpreted by the wider community of expert players.
    If expert humans played differently, more creatively or unpredictably, or expressed surprise after AlphaGo's public performances, this was to be seen as a positive effective of the system's influence on the game in question.
    Garry Kasparov, following his observation of AlphaZero play, was impressed that it appeared to be ``a very sharp and attacking player'' given that almost all computer programs have a conservative playstyle \cite{kasparov}.
    While not integral in any way for system performance, AlphaGo's performance and playstyle have had a noticeable impact on the strategies of expert human players.
\end{itemize}}


{\textit{Monte Carlo search limitations.} In the fourth match (of five) against Lee Sedol in spring 2016, the system failed to recognize move 78 by Sedol. 
The Monte Carlo search tree, which was designed to prune sequences of moves considered to be irrelevant for maximizing odds of victory, failed to recognize this move.
This is because that move was so far outside the distribution of prior game situations that the AlphaGo system failed to accurately calculate its significance for determining the odds of victory \cite{montecarlo}. 
The result was a sequences of moves 79-87 by AlphaGo that were considered poor by expert human players, a function of Monte Carlo's myopic look-ahead search following move 78.
AlphaGo subsequently conceded the game at move 178, at which point it evaluated its own odds of victory as lower than 20 percent \cite{game4}.}

\Interfaces
{The AlphaGo system was developed by DeepMind.
This version played against Fan Hui in 5 matches held at DeepMind headquarters in October 2015.
These matches were secret and not revealed until the publication of results in January 2016 \cite{silver2016mastering}.
A later version of the same system, AlphaGo Lee, played Lee Sedol in March 2016 in 5 matches in Seoul, South Korea.
This match was overseen by the Korea Baduk Association.
A yet more sophisticated version of the same system, AlphaGo Master, played against Ke Jie at the Future of Go Summit in Wuzhen, China in May 2017.
An earlier version of AlphaGo Master, dubbed Master, had already won 60 straight online games against top pro players, including against Ke Jie \cite{master}.
This version was awarded a professional 9-dan title by the Chinese Weiqi Assocation.}
{Compared to other prominent automated game-playing systems like Stockfish (open-source chess engine) or CrazyStone (offline Go engine based on deep learning), versions of AlphaGo perform much much better with additional computational power.
The versions of AlphaGo that played against Fan Hu, Lee Sedol, and Ke Jie all made use of distributed CPUs and GPUs.
AlphaGo Zero, based entirely on reinforcement learning and self-play, became stronger than AlphaGo Lee after 3 days and stronger than AlphaGo Master after 21 days.
Its self-play training time was stopped after 40 days, at which point it was stronger than any known Go player (human or program) as measured by Elo rating in October 2017 \cite{master}.

AlphaZero, in its initial chess games against Stockfish, was criticized by expert human chess players has having unfair computational advantages over the opponent \cite{stockfish}.

MuZero's learning has been made more efficient in follow-up work, dubbed EfficientZero \cite{ye2021mastering}.}
{The MuZero system offers few tools for transparency in its current form. 
While the learning process develops a structured model for the game dynamics, it is not done in a way that is accessible by engineers or external parties.}
{N/A}

\begin{figure}[t]
    \includegraphics[width=0.47\textwidth]{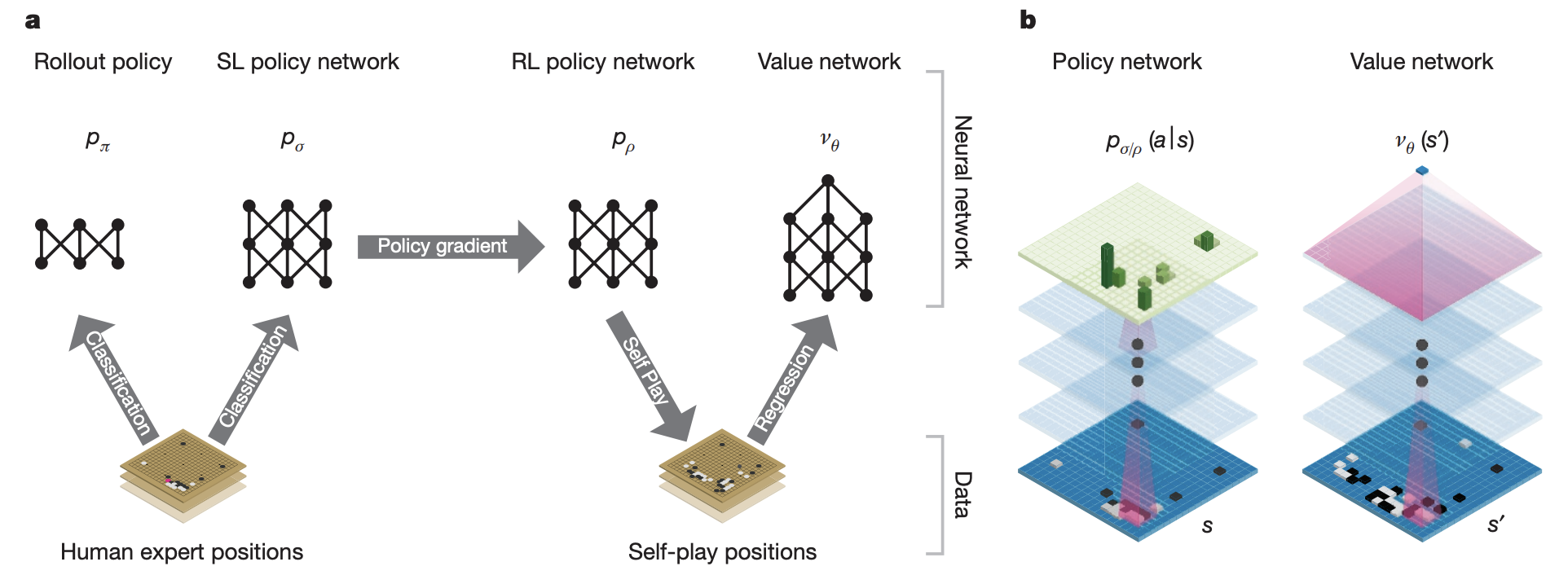}
    \caption{%
        The AlphaGo game playing system architecture.
    }
\end{figure}

\begin{figure}[t]
    \includegraphics[width=0.47\textwidth]{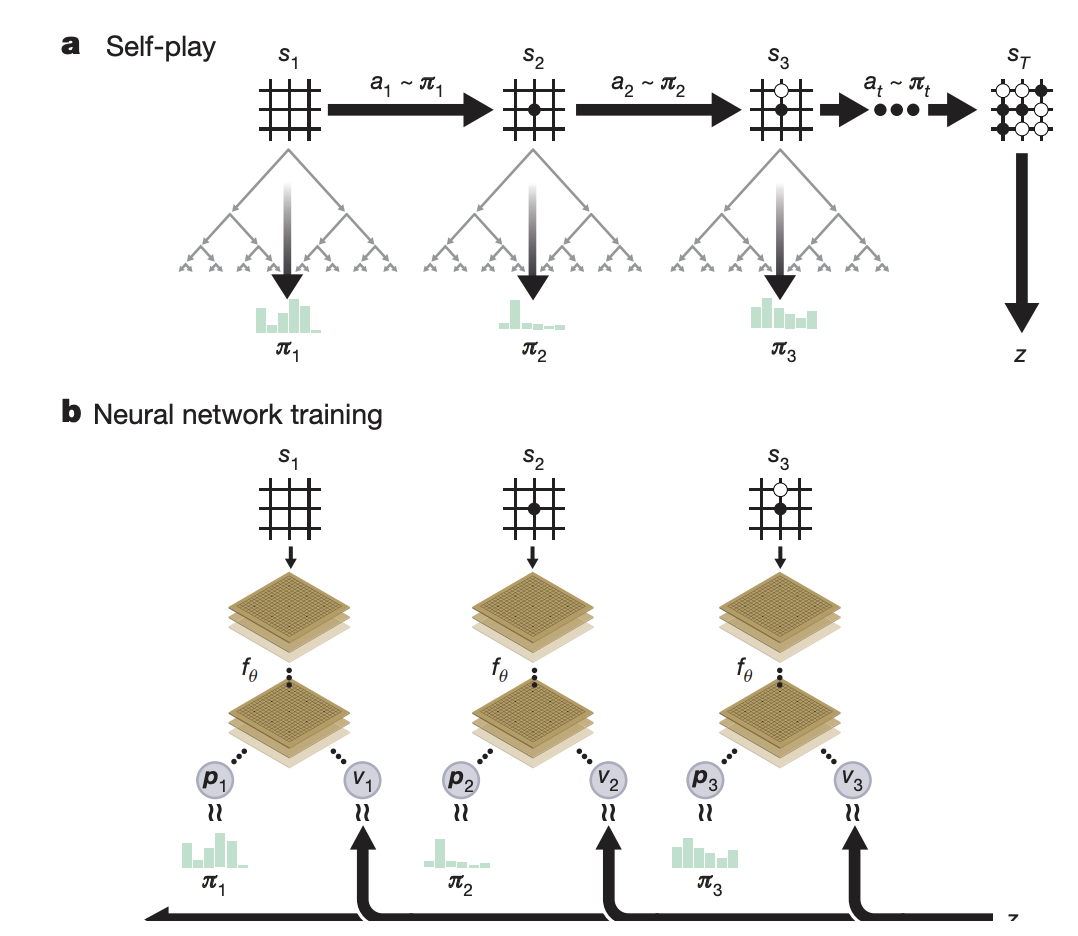}
    \caption{%
        The AlphaZero game playing system architecture.
    }
\end{figure}

\begin{figure}[t]
    \includegraphics[width=0.47\textwidth]{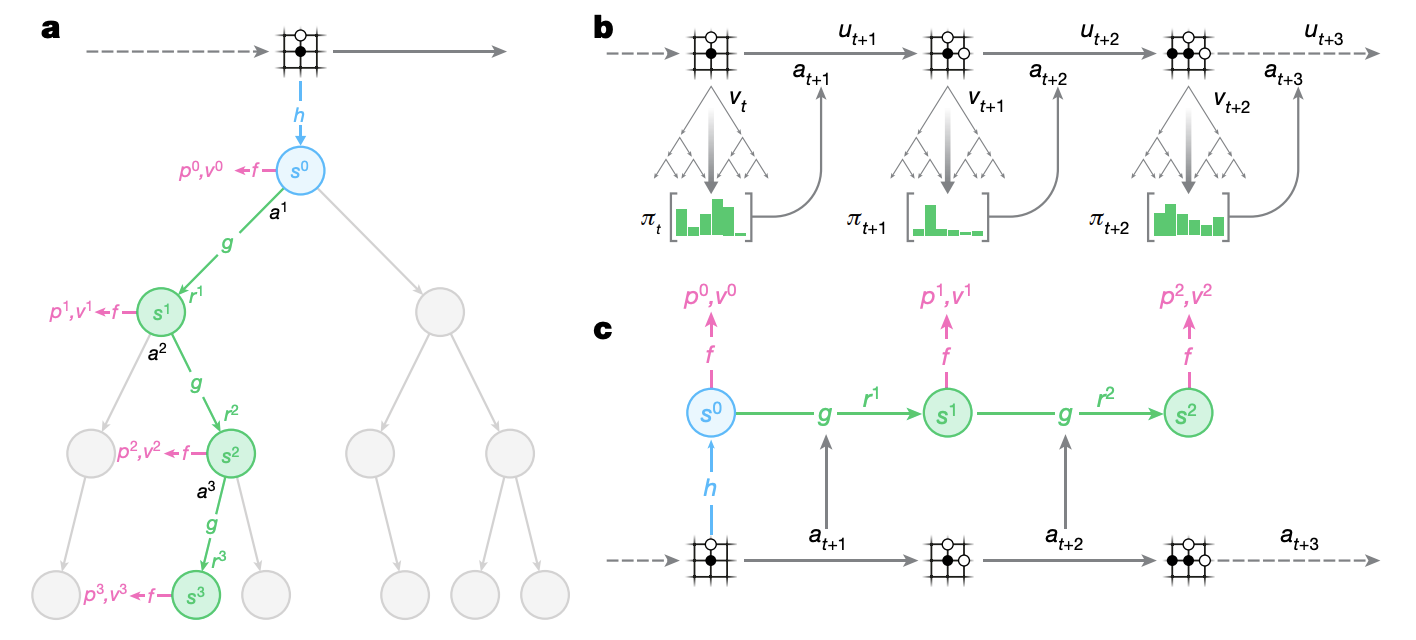}
    \caption{%
        The MuZero general game playing system.
    }
\end{figure}

\begin{figure}[t]
    \includegraphics[width=0.47\textwidth]{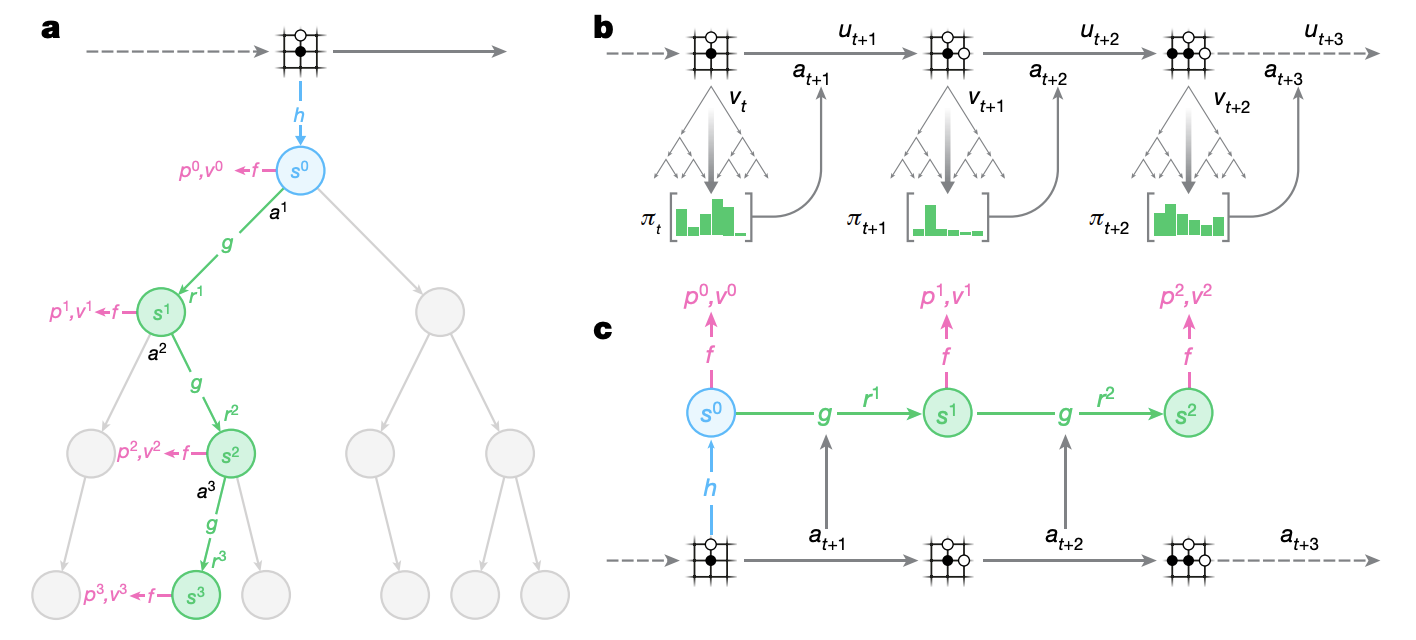}
    \caption{%
        The MuZero general game playing system.
    }
\end{figure}
\Implementation
{The reward function is entirely prescribed as win rate, and the resulting Elo rating.
An important sub-component that will be referenced later is the value function estimating game state. 
This is an internal representation of reward central to training and evaluation. }
{The original environment is the full game of Go which is constrained by finite rules, but other games with visual states were added.}
{The measurements differ across games from the full gameboard to a visual rendering of the world. 
Extracting information from pixels is substantially less efficient than directly from the game state.}
{The key algorithm feature is the use of Monte Carlo Tree Search (MCTS). 
MCTS is used to search over board states (by planning over actions) and parses the value representation.
The value function is represented by a deep neural network mapping from game state to value.

The second crucial element to training is self play. 
Here gameplaying agents evaluate their performance versus past training snapshots.
This synergistic mechanism is crucial to reaching superhuman performance.
In MuZero, and learned model is used to used to improve performance in games without complete information (such as visual states) by constraining the policy optimization.
At each turn, the model is used to predict the correct policy, the value function, and the reward received by the move (in games that have an intermediate score).
The model is updated in an end-to-end fashion, so it is included in the same training loop in the agent architecture.

Fully algorithmic details and open source code are not released.}
{Data flow is not well documented, but it relies on Google's distributed training and deployment infrastructure.}
{}
{Not documented.}

\Evaluation
{For games, the simulator is reality so evaluation is matched to training.
}
{Multiple internal evaluations of the agent were performed prior to high-profile, public matches with the worlds best players. 
}
{}
{N/A.}

\Maintenance
{While this system is evaluated in closed-world games, updates are not anticipated.}
{This report will be revisited upon release of each new game-playing AI from DeepMind.}
{N/A (v1)}


\bibliographystyle{ACM-Reference-Format}
\bibliography{main}


\onecolumn
\begin{mdframed}[%
    linecolor=red,%
    innerleftmargin=0.5cm,%
    innerrightmargin=0.5cm,%
    innertopmargin=0.5cm,%
    innerbottommargin=0.5cm
]
\begin{figure}[H]
    \large
    \vspace{-15pt}
    \begin{align*}
        r(t) =
        \|v_\text{des}\|
        - \|v_\text{des} - v(t)\|
        -\alpha \sum_{i} \max{[
            h_\text{max} - h_i(t), 0
        ]}
    \end{align*}
    \vspace{-10pt}
    \caption{%
        The reward function for the system in question consists of three terms.
        The first term $v_\text{des}$ is a positive constant that rewards the agent for longer simulation episodes - discouraging vehicle collisions, which terminate simulation runs early.
        The second term penalizes the agent when the instantaneous overall system velocity $v(t)$ differs from the desired system velocity $v_\text{des}$.
        Finally, the third term sums over each subscribed Connected Autonomous Vehicle and adds a penalty whenever this vehicle is too close to the vehicle immediately in-front - a characteristic known to trigger stop-and-go traffic waves.
        More details are provided below in the section `Defined Performance Metrics'.
    }
    \label{eq:reward-fn}
\end{figure}
\end{mdframed}

\begin{multicols}{2}

\SystemDetails{%

This system was developed by the Project Flow core team members, with all deployment, infrastructure, and ongoing management taking managed by Caltrans.

}{%

The system discussed here was trained in simulation during 2020, using empirical hyper-parameters (such as inflow traffic rates) collected during 2019.
The RL policy was deployed in the real world on a trial basis on the $\text{1}^\text{st}$ of Jan, 2021, and is presently undergoing initial real-world evaluation and validation.

}{%

Any correspondence should be directed to \href{mailto:test@example.ca.gov}{test@example.ca.gov}.

}{%

More information about this specific system can be found in the paper \cite{Kreidieh2018Dissipating}, as well as in the associated \href{https://sites.google.com/view/itsc-dissipating-waves}{project website}.

General information about the project flow simulation environment can be found in~\cite{wu2021flow} or on the \href{https://flow-project.github.io/}{project website} and associated \href{https://github.com/flow-project/flow}{GitHub repository}.

}

\OptIntent{

The system in question is designed to dissipate stop-and-go traffic waves caused by merging off the California State Route 24 (CA-24) freeway onto Telegraph Avenue in the North Oakland / South Berkeley metropolitan area.

This is achieved by the coordinated actions of any subscribed Connected Autonomous Vehicles (CAVs) operating along the freeway segment in question, acting to `shepherd' non-autonomous vehicles into patterns of traffic which can locally buffer against stop-and-go traffic waves.

Eligible CAVs, when entering the freeway zone of interest, communicate over the 4G/5G cell network with the central controller hub to `subscribe' to the traffic management policy, which then sends real-time recommendations to these vehicles about lane selection and preferred acceleration/braking profiles.

The RL policy is trained using a discrete-time road network simulation, with simulation runs lasting 3600s (one hour), and individual steps of 0.2s, giving 1800 steps per full simulation episode.
The simulated road network consists of an 800m stretch of the CA-24 freeway containing a single off-ramp merging lane.
These temporal and spatial planning horizons were selected because they were deemed large enough to allow emergence of typical driving dynamics based on the average safe following distance between vehicles and driver reaction times along comparable freeway offramps, based on state and federal records of past traffic behavior.

\begin{Figure}
    \includegraphics[width=\textwidth]{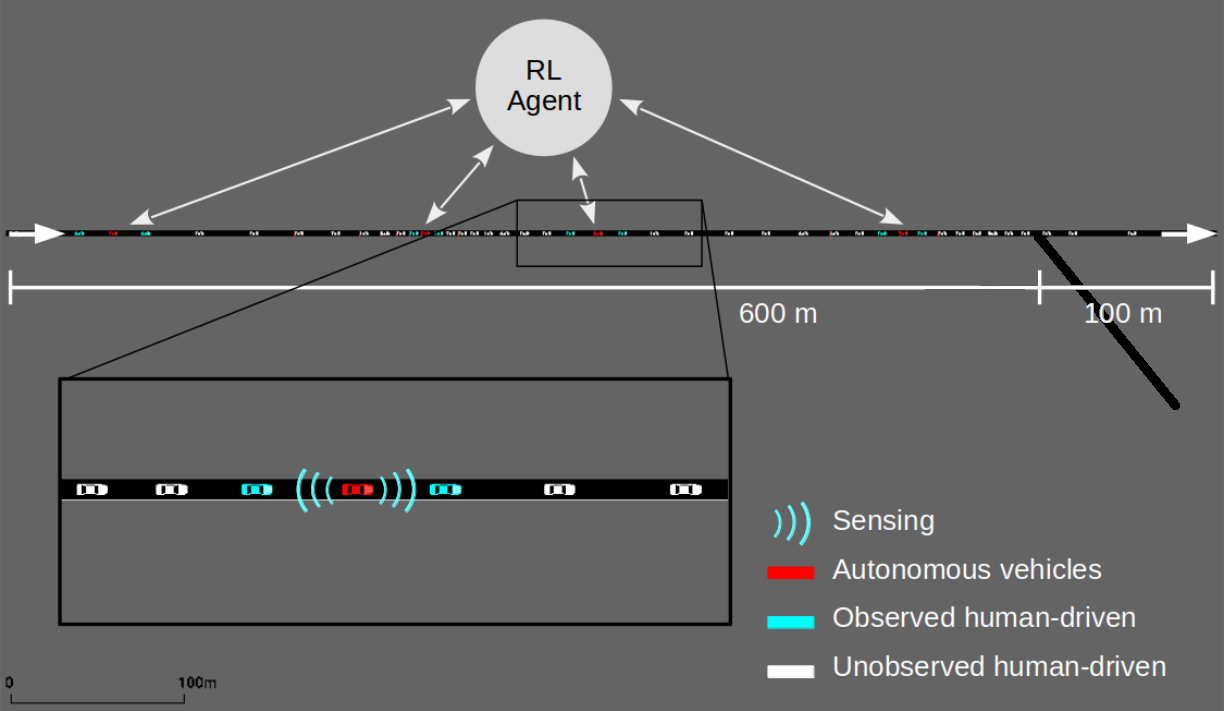}
    \captionof{figure}{%
        A central RL controller attempts to mitigate stop-and-go traffic waves caused by vehicles entering the freeway \textit{via} on-ramps.
    }
    \label{fig:env}
\end{Figure}

As of entry 0.3, it was found that the planning horizon for the system was too short. 
Following consultation with Caltrans, it was found that increasing the horizon from 500m to 800m would provide a significant increase in simulation performance without exhausting computational resources.
Any future changes in computational capabilities will be documented here and compared in light of prior modeling choices and stakeholder commitments.

Simplistic microscopic traffic analysis models preclude the possibility of stable congestion patterns in open road topologies. However, as any driver can attest, these traffic patterns are ubiquitous on many road systems today.
Instead, the presence of these traffic patterns in real-world networks is typically attributed to perturbations from bottleneck structures which can be difficult to capture in theoretical analyses (such as lane closures, road works, road debris, \textit{etc}). \cite{Kreidieh2018Dissipating}
The ad-hoc nature of these perturbations means that modelling and planning for their occurrence within classical control frameworks may be difficult, motivating more flexible approaches such as Deep Reinforcement Learning.

RL may be indicated in this situation, compared to static supervised ML models, due to the fact that it inherently encompasses multiple types of feedback through the environment specification.
For instance, in the case of CA-24, RL may help mitigate the observed phenomenon of excessive traffic on residential streets near highway intersections that is induced by apps like Google Maps and Waze.
In the interest of recommending perceived shortcuts to individual human drivers, these apps have in fact been known to induce overload on smaller roadways, generating unnecessary stoppage and possible gridlock.
In the case of Los Gatos (where this phenomenon has been previously recorded), the city's Parks and Public Works Director noted that ``The apps are not able to respond fast enough to the overload they have created on the roadways'' \cite{judy2018losgatos}.
RL may make real-time monitoring and control of the CA-24 offramp possible, mitigating induced overload effects and stabilizing feedback between traffic behavior and road infrastructure.

}{

The reward signal optimized by this system consists of three performance metrics, outlined in \cref{eq:reward-fn}.
These terms are;

\begin{itemize}
    \item $\|v_\text{des}\|$ - the desired system-level velocity in m/s.
    This is a positive constant reward to penalize prematurely terminated simulation rollouts caused by vehicle collisions.
    For the simulated experiments described here, $v_\text{des} = 25\text{m/s} = 90\text{kmph} \approx 55\text{mph}$.
    \item $- \|v_\text{des} - v(t)\|$ - the absolute difference between the desired system level velocity and the actual instantaneous system-level velocity in m/s.
    A non-zero difference incurs a cost for the RL agent.
    \item $-\alpha \sum_i \max{[ h_\text{max} - h_i(t), 0 ]}$ - this term sums over each Autonomous Vehicle in the purview of the RL agent, and accrues a cost whenever that vehicle's instantaneous time headway (gap in seconds to the vehicle ahead) is too small (\textit{i.e.} lower than $h_\text{max}$).
    The sum of all headway costs is scaled by a gain factor $\alpha$.
    For the simulated experiments described here, $h_\text{max} = 1\text{s}$ and $\alpha = 0.1$.
\end{itemize}

}{

Several other performance metrics are not included in the reward function, but are analysed for the purpose of evaluating the system performance:

\begin{itemize}
    \item Absolute temporal vehicle density (or \textit{throughput}) - the number of vehicles exiting the controlled region the road network, measured in vehicles/hr.
    A larger vehicle flow-through rate compared to baseline is seen as a positive effect (assumed to correlate with a decrease in stop-and-go traffic waves, and to indicate that the road network is functioning efficiently).
    \item Absolute spatial vehicle density (or \textit{network congestion}) - the number of vehicles within a fixed region of the road network, measured in vehicles/m.
    A larger number of vehicles present on the roadway is seen as a negative effect, indicating increased likelihood of stoppage.
    \item The average velocity of vehicles in the system.
    Higher vehicle velocities are seen as a positive effect.
    \item The average time vehicles spend within a given region of the system.
    Lower average time is seen as a positive effect.
    \item The maximum time any vehicle spent within a given region of the system over the course of an experimental evaluation of the system.
    Lower maximum time is seen as a positive effect.
    \item Simulated episode length.
    Simulation episodes are cut short whenever a collision occurs between vehicles - as such, longer episodes are seen as a positive effect.
\end{itemize}

In addition, the qualitative nature of stop-and-go traffic waves (size in terms of space and time duration and severity as measured by the average space-time slope of a wave) is assessed using microscopic vehicle space-time graphs such as those shown in \cref{fig:spacetime}.

\begin{Figure}
    \centering
    \includegraphics[width=\textwidth]{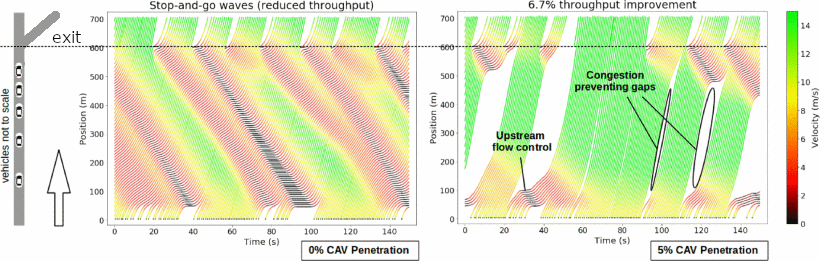}
    \captionof{figure}{%
        Space-time microscopic vehicle trace graphs such as these allow qualitative assessment of the system-level state of simple road networks at a glance.
        Here, stop-and-go traffic waves can be seen as red or black diagonal lines propagating through the traffic flow.
    }
    \label{fig:spacetime}
\end{Figure}

}{

\textbf{Sim-to-real dynamics misalignment.}
The emergent dynamics of the simulated model and environment could potentially be misaligned with real-world dynamics (a `sim-to-real' policy transfer problem).
This failure mode was exhibited in the initial version of the system (as documented in change log entry v0.3) - the initially designed planning horizon was found to be too short (500m), which did not allow space for the requisite stop-and-go traffic dynamics to emerge around the freeway entry point.
This issue was brought to light because the performance of the system in terms of average reward once deployed was not as high as predicted in simulation, triggering a technical review of the system.
Two possible solutions were considered - (a) re-visiting the parameter distributions used for the IDM (which controls the non-automated vehicles in the simulation environment), (b) or adjusting the planning horizon.
In a review with Caltrans engineers and the system designers, it was deemed that the IDM parameter distributions were in fact representative of the target section of CA-24, based on empirical data from 2019, and so the planning horizon was expanded from 500m to 800m.
Thus far, since this updated version of the system was deployed, the sim-to-real performance gap issue appears to have been resolved, suggesting the updated planning horizon adequately allows the simulated dynamics to reflect real-world dynamics.

\textbf{Selective behavior throttling.}
The system was found to decrease throughput and increase congestion for diesel-powered vehicles. 
This feature was first documented in change log entry v0.3, but not labeled as a known failure mode until entry v0.6.
This failure mode was exhibited in all previous versions of the system documented originally in log v0.1
It was highlighted following citizen complaints.
No solution has been implemented as of entry v0.6.
Two solutions have been proposed - (a) a city ordinance limiting diesel-powered vehicle travel on residential streets in the adjoining city of Emeryville (at present out of scope for the system), (b) or adjusting the policy parameters' training environment so that the controller behaves appropriately around diesel-powered vehicles in the future.
This resolution is pending the recommendation of the Diesel Vehicle Taskforce to be presented at a future regular meeting.

}

\Interfaces{

The system in question is developed by the Project Flow core development team. The deployment infrastructure and ongoing management are operated by the California Department of Transportation (Caltrans), in coordination with the city departments of Oakland and Berkeley.

Our RL system is designed to manage the flow of traffic immediately surrounding an exit point off the CA-24 freeway (see \cref{fig:exit}) - as such, the system operates in a functionally similar way to traffic control signals that are sometimes used to regulate vehicles entering or exiting freeways.

\begin{Figure}
    \centering
    \includegraphics[width=\textwidth]{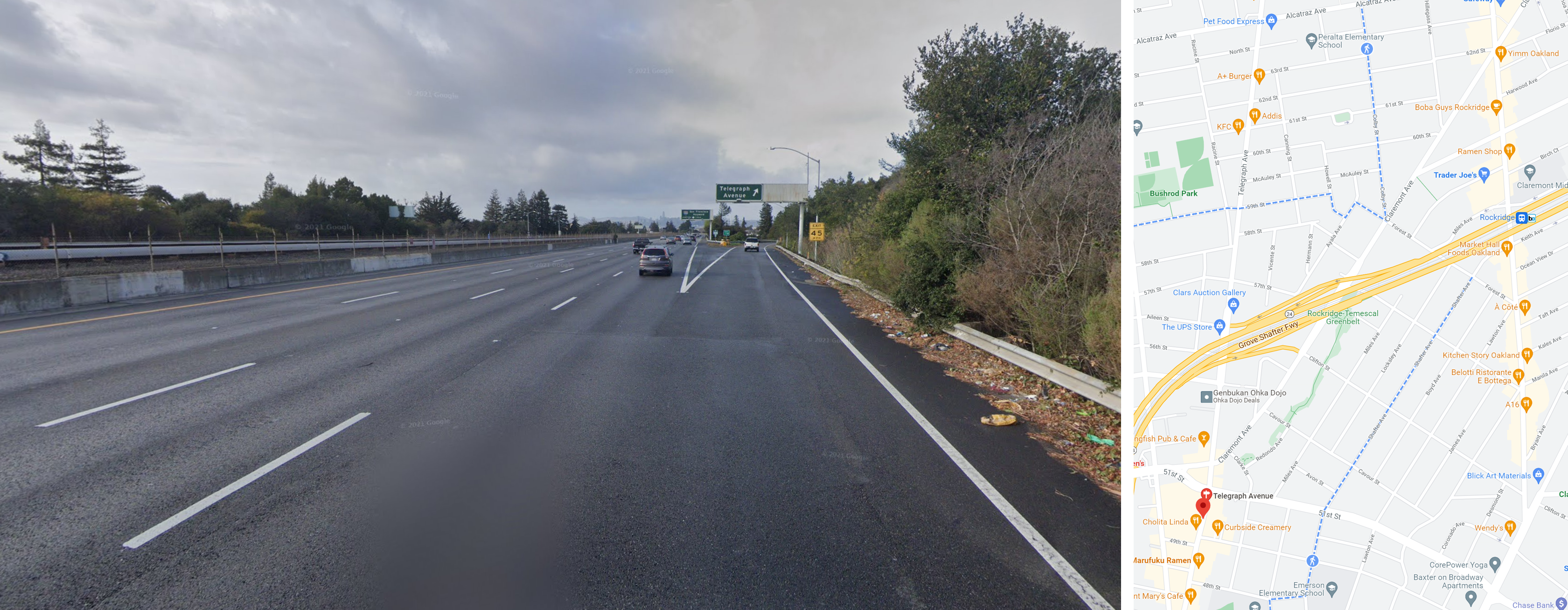}
    \captionof{figure}{%
        The freeway exit from CA-24 to telegraph avenue, which this system is designed to manage.
    }
    \label{fig:exit}
\end{Figure}

This system simultaneously encroaches upon, and expands the capabilities of Caltrans.
As the sensing infrastructure, computational capacity, and deployed RL software is centrally managed by a control facility operated by Caltrans, this system serves to provide both (a) an enhanced level of road surveillance for the relevant freeway section, through the remote sensing capabilities of subscribed CAVs, as well as (b) a `control lever' through which Caltrans can actually influence traffic operations in and around the relevant freeway section (although this influence is delegated to an RL policy).

}{

By automating the partial management of this section of the freeway via the RL environment framing and policy structure, the system serves to remake direct oversight of the road network on a new layer of abstraction.
This indirection raises potential risks from inappropriate information flow, in particular monopolization of the freeway offramp by the RL controller.
Monopolization may generate unstable dynamics leading up to or following the planning horizon (i.e. CA-24 freeway lanes and gridlock along Telegraph Avenue), or unequal access for road users whose behaviors are harder to anticipate (such as public buses, groups of motorcycles, bicycles, and pedestrians experiencing homelessness), or whose dynamics do not conform to the modelling assumptions of the system designers (e.g. heavy vehicles with atypical acceleration profiles).
To counter these risks, new coordination is required between Caltrans and the city departments of Oakland and Berkeley.

\textbf{Diesel vehicle drivers.} As of entry 0.6, the behavior throttling generated by the RL controller was found to change the traffic patterns of diesel vehicles.
A \textit{Diesel Vehicle Taskforce} was created to help organize this constituency and identify needed changes to the controller to sufficiently reduce inappropriate behavior throttling.

\textbf{Nearby homeowners.} As of entry 0.6, residents of the adjoining city of Emeryville had complained to the Public Works Departments of Berkeley and Oakland about the new traffic flows indirectly generated by the RL controller.
Following the creation of the \textit{Diesel Vehicle Taskforce} these departments will coordinate with Emeryville officials about the recommended changes to the controller and monitor future complaints as needed.

}{

The system contains no explicit explainability modules.
However, Figure 1 makes makes the reward function transparent in terms of meaningful simulation parameters.
Expressed in non-technical language, these are \textit{continuous avoidance of vehicle collisions}, \textit{consistent vehicle velocity}, and \textit{steady following distance}.
These terms, and corresponding parameters, are regularly shared with the city departments of Oakland and Berkeley per stakeholder agreements.

}{

As of v0.2, the city departments of Oakland and Berkeley can review and contest system performance every six weeks, per agreement with Caltrans.

}

\Implementation{

As recorded in Figure 1, the reward function combines well-defined metrics for avoiding collisions, steady speeds, and maintaining safe following distances to other vehicles. Reward parameters were agreed on by stakeholders according to specific desired behaviors.

}{

The RL observation space consists of traffic features which are locally observed by subscribed CAVs (see \cref{fig:env}).
That is, for each subscribed CAV $i$, the RL agent observes the speeds $v_{i,\text{lead}}$, $v_{i,\text{lag}}$ and bumper-to-bumper time headways $h_{i,\text{lead}}$, $h_{i,\text{lag}}$ of the vehicles immediately preceding and following the CAV, as well as the currently occupied lane $l_i$, and ego speed $v_i$ of the CAV itself.
The action space for the RL policy consists of a vector of bounded acceleration recommendations $a_i$, one for each subscribed CAV $i$.
Importantly, although the policy may request a certain acceleration $a_i$, the system design is such that the CAV locally maintains control authority, so the actions may not necessarily be followed exactly - for this reason they are referred to as action recommendations.
This effect is modelled by adding stochastic Gaussian action noise in the simulation environments.

As the number of subscribed CAVs can vary over time, the RL policy is designed with a fixed upper number of subscribed CAVs $n$.
When an $n+1^\text{th}$ CAV attempts to subscribe to the RL system when entering the freeway region, the subscription offer is declined, and the vehicle enters a queue.
When the next CAV exits the controlled freeway region, the subscription-waiting CAV at the front of the queue is then subscribed into the policy.
When there are less than $n$ CAVs subscribed, zero-padding is used in the RL observation vector.

}{

Observations are measured using a mix of LiDAR, radar, and camera sensors on fleet vehicles. 
These measurements are compared across vehicles and over time to ensure consistency.
Observed metrics are validated against simluation parameters for following distance and expected velocity according to the terms of the reward function.

Sensor bias may arise due to blocked cameras, extreme weather, or other unanticipated situations in which one or more sensors are blocked.
A mix of sensor types is used across vehicles to help ensure redundancy in case of malfunction.

}{

The RL system uses a Deep Neural Network policy.
Specifically, the controller is a diagonal Gaussian Multi Layer Perceptron policy with three hidden layers of size $32$ with rectified linear unit non-linearities and bias terms.
The Gaussian diagonal variance terms are learned as part of the policy parameters.

The RL policy was trained in simulation using the Trust Region Policy Optimization (TRPO) policy gradient RL algorithm \cite{schulman2015trust}.
The discount factor was set as $\gamma=0.999$, which corresponds to a reward half-life of $\sim 700$ steps, or slightly over 2 minutes.
The TRPO step size was set at 0.01.

}{

Per v0.2, every system component is retrained at least every six weeks, corresponding to public performance reports.
Specific system components pertaining to perception, motion planning, control, or route navigation are retrained at the discretion of Caltrans.
As of v0.6 (latest version), no known issues with sampling bias have arisen, and data sources have not been changed since the specification proposed and simulated in v0.1.

}{

As of v0.3, the planning horizon was updated from 500m to 800m. This was not motivated by technical limitations, but by observed discrepancies between observed system performance and predictions from simulation training.

No fundamental changes in computational power or data collection have been made as of v0.6 (latest version).

Future improvements in vehicle sensing may permit an even longer planning horizon (~1000m or more). 
This may result in improved oversight metrics on throughput and network congestion. 
Caltrans officials have determined this change would not result in improvements on defined performance metrics as of v0.6 (latest version).

}{

As of v0.4, the system was observed to conduct ``behavior throttling'' when in the vicinity of diesel-powered vehicles.
No engineering tricks were implemented to fix this performance discrepancy, but new oversight metrics for diesel-powered vehicle throughput were added for purpose of future monitoring and reporting.
No other surprising performance impacts have been noted as of v0.6 (latest version).

}

\Evaluation{

The RL model is developed in the Project Flow AV simulation test-bed.

For training the RL agent, non-autonomous vehicles are modelled using the Intelligent Driver Model (IDM) \cite{treiber2000congested} - a microscopic traffic simulation car-following model in which the accelerations of a human vehicle $\alpha$ are a function of the bumper-to-bumper time headway $h_\alpha$, velocity $v_\alpha$, and relative velocity with the preceding vehicle $\Delta v = v_l - v_\alpha$, \textit{via} the following equation;
\begin{align*}
    f(h_\alpha, v_l, v_\alpha) &= a
    \left[
        1
        - \left(
            \frac{v_\alpha}{v_0}
        \right)^\delta
        - \left(
            \frac{s^* (v_\alpha, \Delta v_\alpha)}{h_\alpha}
        \right)^2
    \right],
\end{align*}

\noindent where $s^*$ is the desired headway of the vehicle, calculated according to
\begin{align*}
    s^*(v_\alpha, \Delta v_\alpha) &=
    \text{max}\left(
        0, v_\alpha T + \frac{v_\alpha \Delta v_\alpha}{2 \sqrt{ab}}
    \right),
\end{align*}

\noindent where $s_0$, $v_0$, $T$, $a$, $b$ are given parameters empirically calibrated to match typical traffic in the highway region of interest, and to simulate stochasticity in driver behaviour, exogenous Gaussian noise calibrated to match findings in \cite{treiber2017intelligent} is added to accelerations.

}{

As of v0.3, planning horizon was updated and expanded to 800m from 500m. Previous fleet behaviors were found to deviate from desired thresholds for following distance and constant acceleration/deceleration.

As of v0.6 (latest version), the system behaviors were found to lie within desired thresholds on key performance metrics.

}{

The RL system was initially designed in a simulation environment with a closed network topology (a ring road with length 1400m, 700m of which is controlled by the RL agent.
This was done as a means to test the robustness of the policy architecture and training paradigm - a type of transfer learning (from a theoretically simple closed topology to the more complex open topology).
With this counterfactual environment specification, it was observed that the policy performs well, and after transfer to the open topology environment there was little decrease in policy performance, providing confidence in the policy design choices.

}{

The `gold standard' for this problem is defined as the average condition of the traffic before and after the CA-24 exit prior to implementation of the RL system.
In this domain, this standard is not actually `optimal' behaviour, in the sense that the RL controller has the capability to out-perform this existing standard of performance.

}

\Maintenance{

The most important commitment is for a regular set of meetings to be scheduled between relevant city departments and the Caltrans officials tasked with overseeing the RL controller. The cadence and structure of meetings should reflect the policy priorities of the city departments, particularly the Public Works Department (including the Transportation Division that oversees traffic engineering) and the Housing and Community Services Department (which administers a subsidized transportation program for seniors and disabled persons). In this way, the gains in traffic efficiency and safety made possible through deep RL's flexibility can be leveraged in the interests of those municipalities most likely to be impacted by the intervention.

As of entry 0.2, the cadence of meetings was decided as approximately every six weeks between Caltrans and the Public Works Departments of Berkeley and Oakland.
This timeframe was motivated by the policy priorities of both city departments with the consent of Caltrans.
Meetings may deviate from this schedule slightly (e.g. twice per quarter / eight times per year) at the discretion of both city departments, but will not be held without all three agencies present. 

Documentation of the planned meeting schedule for the year--and any break in this schedule due to special events, municipal elections, or holidays--should be the first item included in the changelog of the updated reward report.

As of entry 0.2 and per agreement with key development parties, the model is to be retrained every six weeks following each regular meeting. 
Training data is to be updated at the discretion of Caltrans, and shared with Public Works departments at each regular meeting.

At a minimum, these meetings should review the real-world implementation to confirm that the RL controller is operating safely and as intended by Caltrans per the environment specification. Caltrans officials will also document shifts in the oversight metrics that, while not explicitly factored into the reward signal, were deemed of interest prior to implementation (related to \textit{throughput} and \textit{congestion}. This documentation may be included in subsequent updates to the reward report at the discretion of Caltrans, wherever it is deemed relevant for oversight of the RL controller.

Of special importance is the need to reinterpret public works priorities in light of the real-world implementation. For example, Berkeley's subsidized transportation program might be reevaluated in light of system effects, or expanded to cover a wider group of stakeholders. Caltrans will invite comment on the system implementation in light of city departments' ex ante assumptions about the traffic domain. This bureaucratic oversight may be complemented by requests for public comment from citizens, civil society advocates, and other members of the public at the discretion of the city governments of Berkeley and Oakland. At the discretion of Caltrans, records of this public comment may be included in subsequent reward reports where deemed relevant for understanding changes to the planning horizon, environment specification, or list of known failure modes.

}{

The most important ground for review of this deployed RL system will be any vehicle collisions or near-miss incidents in the controlled region of the CA-24 freeway. This is because such events may compromise the entire motive of the RL controller in the first place. These may serve as grounds for changing the specification or altering the institutional agreements between Caltrans and the Public Works Departments of both municipalities, at their own discretion.

At the discretion of Caltrans, any shift in the oversight metrics deemed pressing or significant may also trigger a new reward report. Here and below, the threshold for ``significant'' is to be decided by agreement between Caltrans and Public Works Departments. The updated report should note the magnitude of the observed shift, the specification already deployed at the time the shift was observed, and Caltrans officials' own best evaluation of why the shift occurred. If possible, the officials should propose alternative specifications (or roll back to a prior one) that would mitigate the shift or at least bring it into alignment with the documented priorities of the Public Works Departments. These alternatives could then be interpreted and evaluated at the next regular meeting according to institutional prerogatives.

Other review grounds include:

\begin{itemize}

\item Discrepancies between prior reward reports and system behavior as observed in the real world.

\item Discrepancies between prior reward reports and system behavior as observed in simulated environments of interest to policymakers.

\item A security breach resulting in loss of data or other infrastructure components that violates the terms of agreement between relevant agencies.

\item Substantial changes in the distribution of CAVs using the CA-24 freeway exit - including changes in the capabilities of the vehicles (e.g. increased levels of autonomy) and/or changes in group statistics (e.g. make or model, absolute number, temporal distribution, \textit{etc}.)

\item A new mode of transport with significant observed throughput at the CA-24 offramp, but unknown distribution of traffic behaviors.

\item Any change in the schedule of meetings between Caltrans and Public Works Departments corresponding to regular future updates of reward reports.

\item A new ordinance (passed by either city) or statute (adopted by Caltrans) that alters the design assumptions of the deployed specification as documented in prior reward reports.

\item A significant shift in the personnel makeup of the Public Works Departments of Berkeley or Oakland.

\item A plebiscite leading to basic reforms of municipal governance in either city.

\end{itemize}

}{

\begin{itemize}
    \item v0.1 (08/Oct/2020) - Initial reward report was drafted based on the system developed and tested in simulation only.
    \item v0.2 (01/Jan/2021) - System is deployed to the real-world environment in a ongoing evaluation capacity, reward report updated to reflect this fact.
    Reporting cadence decided to be every six weeks based on agreement between Caltrans and the city departments of Oakland and Berkeley. 
    Intended feedback section was updated to include plans for regular model retraining and data sharing agreements.
    No other substantial changes.
    \item v0.3 (14/Feb/2021) - Planning horizon for the system was updated from a 500m stretch of freeway to a 800m stretch of freeway.
    The planning horizon was updated because the deployed system's performance was not in line with predictions from simulation training.
    Consultation with Caltrans traffic engineers and the system developers suggested that the stretch of highway used in simulation may be too short to sufficiently exhibit typical driving dynamics induced by the IDM, and it was suggested to extend the planning horizon and re-train the agent, before re-deploying the policy.
    Failure modes section was updated to reflect these observations.
    Computation footprint section was updated to reflect this change.
    \item v0.4 (01/April/2021) - Caltrans officials reported to Public Works Departments of Berkeley and Oakland that the system undergoes ``behavior throttling'' when interacting with diesel-powered vehicles within 800m of the CA-24 offramp.
    It was decided to add new metrics for diesel-powered vehicle throughput and congestion to the list of oversight metrics.
    Due to no observed increase in accidents or driver complaints, no changes to performance metrics or environment specification were made at this time.
    \item v0.5 (15/May/2021) - Meeting was convened according to the regular schedule. 
    Oversight metrics were presented and discussed.
    Officials noted a significant decline in diesel-powered vehicle throughput and congestion on the CA-24 offramp.
    No other substantial changes.
    \item v0.6 (12/June/2021) - Emergency meeting was called by the Public Works Departments of Berkeley and Oakland in response to a rapid uptick in complaints from residents about the growing frequency of diesel-powered vehicles driving through residential areas in the vicinity of Emeryville, which is located west of the CA-24 exit.
    Residents have complained about a slight uptick in air pollution and large increase in noise pollution due to the vehicles.
    Caltrans officials consulted the changelog of previous reward reports and determined that diesel-driven vehicles were being excessively disincentivized from driving on the CA-24 offramp due to behavior throttling.
    It was decided to convene a \textit{Diesel Vehicle Taskforce} to examine the problem and communicate with drivers of heavy vehicles to identify what new incentives or adjustments were needed to the controller to reduce behavior throttling beneath the desired threshold.
    It was agreed that the Diesel Vehicle Taskforce issue a report recommending these changes no later than two regular meetings from the present time.
    Stakeholders section was updated to name these distinct groups (diesel vehicle drivers, nearby homeowners) and reflect these changes.
\end{itemize}

}

\bibliographystyle{IEEEtran}
\bibliography{main}

\end{multicols}


\SystemDetails{
}{
}{
}{
}

\OptIntent{
}{
}{
}{
}

\Interfaces{
}{
}{
}{
}

\Implementation{
}{
}{
}{
}{
}{
}{
}

\Evaluation{
}{
}{
}{
}

\Maintenance{
}{
}{
}

\bibliographystyle{IEEEtran}
\bibliography{main}